\documentclass{article}

\usepackage{arxiv}
\usepackage[dvipsnames]{xcolor}
\usepackage[utf8]{inputenc} % allow utf-8 input
\usepackage[T1]{fontenc}    % use 8-bit T1 fonts
\usepackage[colorlinks=true]{hyperref}
\hypersetup{
colorlinks=true,
linkcolor={red},
urlcolor={blue},
filecolor={black},
citecolor={Plum},
}% hyperlinks
\usepackage{url}            % simple URL typesetting
\usepackage{booktabs}       % professional-quality tables
\usepackage{amsfonts}       % blackboard math symbols
\usepackage{nicefrac}       % compact symbols for 1/2, etc.
\usepackage{microtype}      % microtypography
\usepackage{graphicx}
\usepackage[square, colon, numbers]{natbib}
\usepackage{doi}
\usepackage[labelfont=bf]{caption}
\usepackage{siunitx}
\usepackage{amsmath}
\usepackage{bbm}
\usepackage{tikz}
\usetikzlibrary{shapes.geometric, arrows.meta, patterns, calc}
\usepackage{pgfplots}
\usepgfplotslibrary{groupplots} 
\pgfplotsset{compat=1.18}
\usepackage{xcolor}

\definecolor{color0}{HTML}{D51317}
\definecolor{color1}{HTML}{F39200}
\definecolor{color2}{HTML}{EFD500}
\definecolor{color3}{HTML}{95C11F}
\definecolor{color4}{HTML}{007B3D}
\definecolor{color5}{HTML}{31B7BC}
\definecolor{color6}{HTML}{0094CD}
\definecolor{color7}{HTML}{164194}
\definecolor{color8}{HTML}{6F286A}
\definecolor{bluebell}{rgb}{0.64, 0.64, 0.82}
\definecolor{burgundy}{rgb}{0.5, 0.0, 0.13}
 
\tikzstyle{ensemble} = [rectangle, rounded corners, minimum width = 2cm, minimum height = 1cm, text centered, draw=black, fill=bluebell!30]
\tikzstyle{model} = [ellipse, minimum width = 2cm,  minimum height = 1cm, text centered, draw=black, fill=burgundy!30]
\tikzstyle{arrow}= [thick, ->, >=stealth]
\tikzstyle{arrow_dashed}= [-,dashed, >=stealth]
\tikzstyle{details_w2p} = [rectangle, rounded corners, minimum width = 10cm, minimum height = 4cm, text centered, draw = black, dashed]
\tikzstyle{details_pp} = [rectangle, rounded corners, minimum width = 10cm, minimum height = 7.25cm, text centered, draw = black, dashed]

\newsavebox{\fakeensembleraw}
\savebox{\fakeensembleraw}{
    \resizebox{2.5cm}{1.5cm}{
    \begin{tikzpicture}
    \draw[arrows = {-Stealth[inset=0pt, length=4pt, angle'=30]}] (0,0) -- (7,0) node[below] {$t$};
    \draw[arrows = {-Stealth[inset=0pt, length=4pt, angle'=30]}] (0,0) -- (0,3) node[left] {$x_t$};
    \draw (0,0) -- (0,-0.1); % Tick mark
    \node at (0,-0.25) {\tiny Init}; % Init
    \draw (5,0) -- (5,-0.1); % Tick mark
    \node at (5.37,-0.25) {\tiny Init+Lead Time}; % Init

    \draw[thick, gray!50, smooth, tension=0.7]
    plot coordinates {(0,1) (1,1.5) (2,0.3) (3,1.2) (4,0.8) (5,1.5)};
    \draw[thick, gray!50, smooth, tension=0.7]
    plot coordinates {(0,1.2) (1,0.7) (2,1.5) (3,0.8) (4, 1.7) (5,0.6)};
    \draw[thick, gray!50, smooth, tension=0.7]
    plot coordinates {(0,1.3) (1,1.7) (2,2.6) (3,1.4) (4, 1.9) (5,1.8)};
    \draw[thick, gray!50, smooth, tension=0.7]
    plot coordinates {(0,0.9) (1,0.6) (2,1.7) (3,1.2) (4, 1) (5,1.3)};
    \draw[thick, gray!50, smooth, tension=0.7]
    plot coordinates {(0,1.1) (1,1.3) (2,2.1) (3,0.9) (4, 1.2) (5,0.8)};
    
    \draw[thick, black]
    plot coordinates {(5,0) (5,1) (5,2) (5,2.3)};
     \fill[gray!30]                
    (5,0)                        % start at x-axis
    plot[smooth] coordinates {(5,0) (5.1, 1.1) (5.5,1.45) (5.1,1.7) (5,2.1)} 
    -- (5,2.3) -- cycle;           % close shape back to x-axis
    \draw[thick, black, smooth, tension=0.7]
    plot coordinates {(5,0) (5.1, 1.1) (5.5,1.45) (5.1,1.7) (5,2.1)};
    \end{tikzpicture}
    }
}

\newsavebox{\fakeensembleraww}
\savebox{\fakeensembleraww}{
    \resizebox{2.5cm}{1.5cm}{
    \begin{tikzpicture}
    \draw[arrows = {-Stealth[inset=0pt, length=4pt, angle'=30]}] (0,0) -- (7,0) node[below] {$t$};
    \draw[arrows = {-Stealth[inset=0pt, length=4pt, angle'=30]}] (0,0) -- (0,3) node[left] {$x_t$};
    \draw (0,0) -- (0,-0.1); % Tick mark
    \node at (0,-0.25) {\tiny Init}; % Init
    \draw (5,0) -- (5,-0.1); % Tick mark
    \node at (5.37,-0.25) {\tiny Init+Lead Time}; % Init

    \draw[thick, gray!50, smooth, tension=0.7]
    plot coordinates {(0,1) (1,1.8) (2,0.7) (3,1.6) (4,1.2) (5,1.1)};
    \draw[thick, gray!50, smooth, tension=0.7]
    plot coordinates {(0,1.3) (1,2.4) (2,1.) (3,1.9) (4, 1.1) (5,1)};
    \draw[thick, gray!50, smooth, tension=0.7]
    plot coordinates {(0,1.2) (1,0.8) (2,2.3) (3,1.3) (4, 2) (5,0.6)};
    \draw[thick, gray!50, smooth, tension=0.7]
    plot coordinates {(0,1.1) (1,0.3) (2,1.8) (3,0.7) (4, 1.9) (5,0.8)};
    \draw[thick, gray!50, smooth, tension=0.7]
    plot coordinates {(0,0.9) (1,1.3) (2,0.3) (3,1.7) (4, 0.9) (5,1.2)};
    
    \draw[thick, black]
    plot coordinates {(5,0) (5,1) (5,2) (5,2.3)};
    \fill[gray!30]                
    (5,0)                        % start at x-axis
    plot[smooth] coordinates {(5,0) (5.1, 0.7) (5.5,1) (5.1,1.3) (5,2)} 
    -- (5,2.3) -- cycle;           % close shape back to x-axis
    \draw[thick, black, smooth, tension=0.7]
    plot coordinates {(5,0) (5.1, 0.7) (5.5,1) (5.1,1.3) (5,2)};
    \end{tikzpicture}
    }
}

\newsavebox{\fakeensemblepp}
\savebox{\fakeensemblepp}{
    \resizebox{2.5cm}{1.5cm}{
    \begin{tikzpicture}
    \draw[arrows = {-Stealth[inset=0pt, length=4pt, angle'=30]}] (0,0) -- (7,0) node[below] {$t$};
    \draw[arrows = {-Stealth[inset=0pt, length=4pt, angle'=30]}] (0,0) -- (0,3) node[left] {$x_t$};
    \draw (0,0) -- (0,-0.1); % Tick mark
    \node at (0,-0.25) {\tiny Init}; % Init
    \draw (5,0) -- (5,-0.1); % Tick mark
    \node at (5.37,-0.25) {\tiny Init+Lead Time}; % Init

    \draw[thick, gray!50, smooth, tension=0.7]
    plot coordinates {(0,1) (1,1.8) (2,0.7) (3,1.6) (4,1.2) (5,1.4)};
    \draw[thick, gray!50, smooth, tension=0.7]
    plot coordinates {(0,1.3) (1,2.1) (2,1.6) (3,1.9) (4, 1.1) (5,1.6)};
    \draw[thick, gray!50, smooth, tension=0.7]
    plot coordinates {(0,0.9) (1,1.2) (2,0.4) (3,1.7) (4, 0.9) (5,1.5)};
    \draw[thick, gray!50, smooth, tension=0.7]
    plot coordinates {(0,1.1) (1,0.3) (2,1.8) (3,0.7) (4, 1.9) (5,1.4)};
    \draw[thick, gray!50, smooth, tension=0.7]
    plot coordinates {(0,1.2) (1,0.6) (2,2.1) (3,1.5) (4, 2.4) (5,1.55)};
    
    \draw[thick, black]
    plot coordinates {(5,0) (5,1) (5,2) (5,2.3)};
    \fill[gray!30]                
    (5,0)                        % start at x-axis
    plot[smooth] coordinates {(5,0) (5.1,0.4) (5.25,0.8) (5.5,1.25) (5.25,1.7) (5,2.3)} 
    -- (5,2.3) -- cycle;           % close shape back to x-axis
    \draw[thick, black, smooth, tension=0.7]
    plot coordinates {(5,0) (5.1,0.4) (5.25,0.8) (5.5,1.25) (5.25,1.7) (5,2.3)};
    \end{tikzpicture}
    }
}

\title{Achieving Skilled and Reliable Daily Probabilistic Forecasts of Wind Power at Subseasonal-to-Seasonal Timescales over France}

\date{} 					% Or removing it

\author{\href{https://orcid.org/0009-0003-3947-2101}{\includegraphics[scale=0.06]{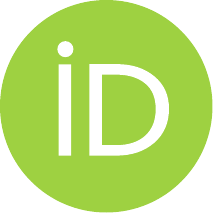}\hspace{1mm}Eloi LINDAS}\textsuperscript{1,3}\thanks{Corresponding author : \href{mailto:eloi.lindas@lsce.ipsl.fr}{eloi.lindas@lsce.ipsl.fr}}\hspace{3cm} Yannig GOUDE\textsuperscript{2,4} \hspace{3cm} Philippe CIAIS\textsuperscript{3}\\
\\
\textsuperscript{1}Atos Inno'Lab TS Bezons, Bezons, France\\
\textsuperscript{2}EDF R\&D Lab, EDF, Palaiseau, France\\
\textsuperscript{3}Laboratoire des Sciences du Climat \& de l'Environnement, CEA, CNRS, UVSQ, \\
Université Paris-Saclay, Gif-sur-Yvette, France\\
\textsuperscript{4}Laboratoire de Mathématiques d'Orsay, CNRS, Université Paris-Saclay, Orsay, France}
\renewcommand{\headeright}{Preprint}
\renewcommand{\undertitle}{Preprint}
\renewcommand{\shorttitle}{Daily S2S Probabilistic Forecasts of French Wind Power}

\hypersetup{
pdftitle={Paper on S2S RES Forecasts},
pdfsubject={cs.LG, stat.ML},
pdfauthor={Eloi LINDAS, Yannig GOUDE, Philippe CIAIS},
pdfkeywords={climate, electricity supply, probabilistic forecasting, machine learning, renewables},
}

\begin{document}

\maketitle
\setcounter{footnote}{0}
\begin{abstract}
In a growing renewable based energy system, accurate and reliable wind power forecasts are crucial for grid stability, balancing supply and demand and market risk management. Even though short-term weather forecasts have been thoroughly used to provide up to $3$ days ahead renewable power predictions, forecasts involving prediction horizons longer than a week still need investigations. Despite the recent progress in subseasonal-to-seasonal weather probabilistic forecasting, their use for wind power prediction usually involves both temporal and spatial aggregation to achieve reasonable skill. In this study, we present a lead time and numerical weather model agnostic forecasting pipeline which enables to transform ECMWF subseasonal-to-seasonal weather forecasts into wind power forecasts for France for lead times ranging from $1$ day to $46$ days at daily resolution. By leveraging a post-processing step of the resulting power ensembles we show that these forecasts improve the climatological baseline by \si{15\percent} to \si{5\percent} for the Continuous Ranked Probability Score and \si{20\percent} to \si{5\percent} for ensemble Mean Squared Error up to $16$ days in advance, before converging towards the climatological skill. This improvement in skill is jointly obtained with near perfect calibration of the forecasts for every lead time. The results suggest that electricity market players could benefit from the extended forecast range up to two weeks to improve their decision making on renewable supply.
\end{abstract}

% keywords can be removed
\keywords{forecasting \and renewable energy \and statistical post-processing \and subseasonal-to-seasonal \and weather}
\paragraph{Data Availability} The raw and post-processed forecasts as well as the baselines, developed in this study can be accessed : \href{https://doi.org/10.5281/zenodo.17311709}{https://doi.org/10.5281/zenodo.17311709}

\section{Introduction}
A sustainable energy system required to mitigate climate change benefits from renewable sources such as wind power to replace traditional fossil-fuel-based power plants. However, compared to these conventional sources, renewables lack flexibility and face temporal as well as spatial intermittency challenges driven by their weather dependency \cite{engeland_2017, gowrisankaran_intermittency_2011, pommeret_optimal_2022}. Not only are renewable power forecasts needed to ensure grid stability by balancing supply and demand, but they are also utilized for Operations and Maintenance (O\&M) scheduling or trading purposes. Accurate and reliable forecasts are thus used and essential to a variety of market players from Transmission System Operator (TSO), electricity suppliers to power traders \cite{amin_2024, peterssen_2024}.

To ensure smooth operations over short-term horizons, the use of weather forecasts has become widespread to provide power forecasts \cite{kim_daily_2017, ledmaoui_forecasting_2023, sharma_predicting_2011}. Modelers leveraging Numerical Weather Predictions (NWP) to satellite images or in-situ observations enabled accurate predictions from a few hours to a few days in advance \cite{juncklaus_2022, wang_review_2019}. On the longer timescales, climate projections and seasonal outlooks were used to determine the potential of renewables, the change in variability induced by climate change, or to anticipate a seasonal anomaly \cite{bloomfield_meteorological_2020, solaun_2019, van_der_wiel_influence_2019}. Long-term energy management highly benefits from such information, reducing the risk incurred by the electricity system transitioning towards intermittent power sources.

However, subseasonal-to-seasonal (S2S) forecasts, which lie between short-term weather forecasts and long-term climate outlooks, are still undervalued by practitioners \cite{white_review_2017}. From better O\&M and grid management to stable price predictions and market risk mitigation, having accurate power forecasts up to a month is crucial for a clean energy grid \cite{dorrington_2020, orlov_better_2020}. Despite the growing interest of researchers over the past decades, leading to an increase in their predictive skill, attention to S2S power forecasts remains small compared to short-term forecasts.

In fact, most of the studies focus on the weather predictability at such horizons and its link with climate patterns such as the North Atlantic Oscillation or El Niño \cite{besteiro_2022, beerli_2017, lynch_verification_2014, soret_sub_seasonal_2019}. Another round of research papers try to improve the forecast quality of key weather variables for power generation such as wind speeds or solar irradiance \cite{chinta_2023, mouatadid_subseasonalclimateusa_nodate, sinha2021subseasonal, soret_sub_seasonal_2019}. While a better understanding and improvements in predictability at such timescales are crucial, converting the weather into proper power forecasts requires another step involving dedicated mid-term power models. Knowing the next low wind speed window to schedule turbine maintenance is important, but it could be improved by also having a generation forecast. Often, a simple physical or linear model is used to transform weather variables into energy or capacity factor, falling short of grasping the entire relationship and its nuances~\cite{bett_simplified_2022, bloomfield_sub-seasonal_2021, lledo_seasonal_2019}.

Furthermore, predictability at S2S horizons is harder to unravel. Since such forecasts are better suited to answer whether the next month is going to be colder or hotter than average, many meteorologists advocate for temporal aggregation to get skilled forecasts and draw quantitative conclusions \cite{shaowu}. Even with the increase in predictive skill at extended range, accurate subseasonal-to-seasonal forecasts at hourly granularity are still not in reach due to their high uncertainty induced by the chaotic nature of the system. Hence why, many research works, focusing on energy, used a weekly or monthly aggregation to show that predictability can arise when using such forecasts, provided that the data is averaged over a long period of time \cite{bett_simplified_2022, bloomfield_sub-seasonal_2021, soret_sub_seasonal_2019}. To the best of the authors' knowledge, no study showed skill at temporal resolution higher than a week. In addition, S2S forecasts can also be aggregated spatially to derive information for a region or a country because their spatial resolution is coarser than short-term forecasts. This aggregation step also helps to improve the skill and reliability of the predictions by smoothing out the spatial forecast errors.

Another key aspect of S2S forecasts is their calibration and reliability. Indeed, most modern weather services provide them as probabilistic forecasts using an Ensemble Prediction System (EPS). An ensemble produces more information than a deterministic prediction as it offers uncertainty quantification through its spread. Still, it has been shown that EPS suffers from biased and under-dispersive predictions \cite{sperati_application_2016, vannitsem_statistical_2018}. This means that the ensemble spread does not describe the uncertainty correctly, usually by being too narrow. To compensate for these aspects, statistical post-processing models have been developed to correct the raw forecasts \cite{vannitsem_statistical_2021}.

Although a range of post-processing techniques for ensemble weather forecasts have been successfully applied in both research and operational contexts, only a few works have tried to post-process weather S2S data \cite{horat_deep_2024, ryu_2025}. The literature is scarcer for ensemble power forecasts and so S2S probabilistic power forecasts. Yet, this post-processing step has the potential to bridge the gap between imprecise and unreliable forecasts and fully calibrated and skilled predictions. Using tools covering parametric models, such as model output statistics \cite{gneiting_probabilistic_2014, gneiting_calibrated_2005}, to machine learning based methods like quantile regression or neural networks \cite{horat_improving_2025, rasp_neural_2018, taillardat_calibrated_2016}, practitioners can correct the bias and spread of the raw forecasts and achieve reliable forecasts.

Therefore, this paper focuses on trying to provide informative S2S probabilistic wind power forecasts over France without using temporal nor spatial aggregation. It addresses the following research questions: Can S2S wind power forecasts show skill at daily resolution up to 6 weeks in advance ? Can we achieve calibrated ensemble predictions and improve the forecasting skill by making use of a post-processing step ? The forecasting pipeline, consisting of S2S weather data, a weather to power model that ingest entire weather fields without prior aggregation, and a post-processing step applied onto the power forecasts, is described in section \ref{sec:data_methods_section}. Then, section \ref{sec:results_section} assesses the skill as well as the calibration of the daily wind power forecasts over lead times ranging from 1 to 46 days that were initialized in 2023-2024. The findings of this study are then summarized in section \ref{sec:conclusion}.

\begin{figure}[ht!]
  \centering
  \input{Scheme/Tex/workflow}
  \caption{Forecasting pipeline of the study represented schematically. S2S weather ensembles are converted to S2S Wind Power forecasts before being post-processed to account for the bias and under dispersion. The details of the weather-to-power model and post-processing model are presented in the dashed boxes. The weather-to-power model leverage spatially explicit weather data as well as location and power capacity of wind farms in France trough a Convolutionnal Neural Network. The post-processing model is trained following an online procedure. However, available data for training is limited by the observation availability gap with the forecast horizon. As represented on the temporal axis, since forecasts are issued at an initialization date with a forecast horizon, the corresponding observation (ground truth) is not available at the same time leading to a temporal gap between training and inference.}
  \label{fig:flowchart}
\end{figure}

\section{Data \& Methods}\label{sec:data_methods_section}
The following section presents the subseasonal-to-seasonal weather forecasts data used (section \ref{sec:s2s_data}) as well as the methodology to convert these weather forecasts into wind power forecasts (section \ref{sec:weather2power_model}) which is also presented schematically in the flowchart of figure \ref{fig:flowchart}. The baselines and metrics used to assess the skill of the forecasts are presented in section \ref{sec:baselines} before presenting the post-processing techniques used to correct the forecasts biases and under-dispersion to improve the forecasts quality (section \ref{sec:pp_methods}).

\newpage
\subsection{S2S Forecast Data}\label{sec:s2s_data}
Subseasonal-to-Seasonal (S2S) forecasts from the European Center for Medium-Range Weather Forecasts (ECMWF) and several other National Weather Services (NWS) are available in the S2S database\footnote{This database is available on the ECMWF website at :\url{https://www.ecmwf.int/en/forecasts/dataset/sub-seasonal-seasonal-prediction}. Accessed: 11\textsuperscript{th} June 2025} \cite{s2s_database}. This database contains both operational forecasts and reforecasts. We retrieved the forecasts from the ECMWF itself, which were initialized in 2023 and 2024. The data covers lead times ranging from \si{1} day to \si{46} days with a spatial resolution of \si{1,5\degree}$\times$\si{1,5\degree}. Depending on the weather variable at hand, the temporal resolution can be \si{3} hourly, \si{6} hourly, or daily. Each variable was temporally aggregated to the daily resolution, as it's the smallest common resolution and the highest at which predictability can arise at longer lead times. To match the spatial resolution used by our Weather to Power model (section \ref{sec:weather2power_model}), the data were interpolated from \si{1,5\degree}$\times$\si{1,5\degree} to \si{0,25\degree}$\times$\si{0,25\degree} using the MIR interpolation framework \cite{maciel_2017}. The forecasts ensemble comprised \si{50} perturbed members plus \si{1} control member initialized every Monday and Thursday up until the 23\textsuperscript{th} June 2023. After this date, the ensemble prediction system switched towards daily initialization and \si{100} perturbed members plus \si{1} control member. To be consistent over the period, we retrieved all initializations available for all lead times over the years 2023-2024, but we kept a fixed ensemble size of \si{51} members. We were able to do so because the ensemble members are equally probable. This provided forecasts covering dates ranging from 2023 to early 2025.

As this study is focusing on France, the following bounding box was used [\si{-4,55\degree}; \si{42,5\degree}; \si{7,95\degree}; \si{51\degree}] to download the data before using the country boundaries to crop the weather fields. Wind speeds at \si{10~\meter} and \si{100~\meter} were used for wind power conversion. The S2S database only provides \si{10~\meter} eastward and northward wind speeds. Thus, these were extrapolated to \si{100~\meter} using a wind shear law:
\begin{equation}
    u_{100\mathrm{m}} = u_{10\mathrm{m}}\left(\dfrac{z_{100m}}{z_{10m}} \right)^\alpha
\end{equation}
with $u$ the wind speed, $z$ the height from the ground and $\alpha = \dfrac{1}{7}$ as commonly done in the wind modeling community \cite{bloomfield_sub-seasonal_2021, gualtieri_methods_2012, justus_height_1976}.

\subsection{Weather to Power Model}\label{sec:weather2power_model}
The modelling pipeline used to transform weather data into national wind power predictions follows the work from Lindas \textit{et al.}. It is represented in the upper dashed box of figure \ref{fig:flowchart} and briefly described here \cite{lindas_2025}. The data-driven methodology is two-staged. First, the weather fields are weighted by the power capacity in each grid cell, and each time step using the following weights:
\begin{equation}\label{eq:capacity_weighting}
    \omega_{i,j} = \dfrac{P_{i,j}(t)}{\sum_{i,j,t}P_{i,j}(t)}
\end{equation}
where $P$ is the wind power capacity available in grid cell $i,j$ at time $t$. This enables to focus on grid cells with producing wind farms while also taking into account the increase in installed power capacity over time .The power capacity is obtained through the ORE power plant inventory\footnote{This inventory can be found on the ORE website: \href{https://opendata.agenceore.fr/explore/dataset/registre-national-installation-production-stockage-electricite-agrege/information/?stage_theme=true&disjunctive.departement&disjunctive.region&disjunctive.epci&disjunctive.filiere&disjunctive.technologie&disjunctive.combustiblessecondaires&disjunctive.combustible&disjunctive.gestionnaire&disjunctive.regime}{https://opendata.agenceore.fr/explore/dataset}. Accessed 24\textsuperscript{th} June 2025.}, which gives the power capacity, starting date, and approximate location of power plants implemented in France. Then, a machine learning model trained to predict the daily French wind power supply from the power-weighted weather data is applied. For its training, our weather-to-power model utilizes data from the ERA5 reanalysis at \si{0,25\degree}$\times$\si{0,25\degree} resolution \cite{era5} as inputs, while the entire French wind power supply sourced from the TSO database\footnote{All the data regarding France's electricity consumption and generation can be found in RTE eCO\textsubscript{2}mix: \url{https://www.rte-france.com/en/eco2mix}. Accessed: 24\textsuperscript{th} June 2025.} is used as the training target. The machine learning model architecture was decided based on the benchmark results of a previous study \cite{lindas_2025}. Namely, we used a Convolutionnal Neural Network (CNN) architecture, which is able to ingest weighted weather fields as inputs, therefore avoiding the need for any spatial aggregation and leveraging the entire geospatial information.

All the data were aggregated to a daily resolution, and the machine learning model was trained to predict a day's wind power supply using the same day's power-weighted weather data. The model was trained on data covering the period 2012-2022 before being used on the S2S weather forecasts to provide S2S wind power forecasts. The S2S forecasts were also power-weighted prior to being fed to the model (see equation \ref{eq:capacity_weighting}). This technique has two main advantages. First, the model is lead time agnostic, meaning that we can use forecast data with any horizon, and power forecasts will be obtained. The only constraint to retrieve a forecast for a given date is to feed it with the weather conditions of that date. It allows to avoid having multiple models for multiple horizons. Second, the model is not required to be trained on S2S forecast data, completely bypassing the need for reforecasts data and retraining whenever there is an update to the numerical weather prediction model. Last, it also allows to use any weather forecasting model as long as the parameters and the spatial resolution are the same as the ones the model was trained on.

\subsection{Comparison Baselines \& Metrics}\label{sec:baselines}
To assess if the subseasonal-to-seasonal forecasts are informative and to check if post-processing actually improves their performance, two baselines were considered for comparison: a climatological baseline and a bootstrap climatological baseline. Traditionally, S2S weather forecasts are compared with climatological forecast as it is the standard way to retrieve prediction at such timescales.

That's why, using ERA5 data, a weather climatology over the last \si{30} years was computed and converted to climatological power forecasts using our weather-to-power model. An ensemble composed of data for the same date using the last \si{30} years of data is constructed before being weighted using the power capacity of 2023-2024 to match the forecasts. This will be referred to as the climatological baseline in the following sections. In addition to the climatological baseline based on a weather climatological ensemble, we built a climatological bootstrap baseline to consider more diverse weather conditions in the ensemble. Instead of concatenating the weather for a given date over \si{30} years, we used a window around that date to keep similar weather conditions for each year before drawing samples with replacement to build an ensemble of past weather conditions for the given date. This allow to increase the representativeness of weather conditions. Using a correlogram analysis for different weather variables, a window size of 3 days centered around the date was used. This led to 90 possible weather states that could be drawn from. To be consistent with the S2S data, an ensemble with 51 draws was constructed.

In order to compare the skill of the S2S forecasts to our baseline, a skill score with regard to the climatological baseline was computed:
\begin{equation}
    SS = \dfrac{\mathrm{Skill_{S2S} - \mathrm{Skill_{clim}}}}{\mathrm{Skill_{perfect}} - \mathrm{Skill_{clim}}}
\end{equation}
where $\mathrm{Skill}$ is a metric of interest and $\mathrm{Skill_{perfect}}$ is the skill of a perfect forecast. In our case, we considered the following metrics : the Continuous Ranked Probability Score (CRPS) \cite{gneiting_strictly_2007} and the Mean Squared Error (MSE), Root Mean Squared Error (RMSE), Mean Absolute Error (MAE) and $\mathrm{R}^2$ score (R2) of the ensemble mean. The CRPS is a probabilistic metric taking into account the entire distribution of the ensemble. It is defined as:
\begin{equation}\label{eq:CRPS}
    CRPS(F(\hat{y}),y) = \int \left(F(\hat{y}) - \mathbbm{1}_{x-y}\right)^2 dx 
\end{equation}
with $\hat{y}$ the predicted ensemble, $F(\hat{y})$ it's cumulative distribution function (CDF), $y$ the observation and $\mathbbm{1}$ denoting an indicator function. The CRPS is a measure for calibration and sharpness of the predicted ensemble distribution. It compares the ensemble CDF with the observation CDF which is a single point. The lower the CRPS the better. Since it is defined for a single sample, we can average it over a period of $n$ samples such that $CRPS = \sum_{i=1}^n CRPS_i$.\break
The MSE, RMSE, MAE and R2 of the ensemble mean are deterministic metrics defined as :
\begin{equation}\label{eq:deterministic_metrics}
  \begin{aligned}
    MSE_{Ens} &= \dfrac{1}{n}\sum_i (\bar{\hat{y}}_i-y_i)^2  & MAE_{Ens} &= \dfrac{1}{n}\sum_i |\bar{\hat{y}}_i-y_i| & R^2 = 1 - \dfrac{\sum_i (\bar{\hat{y}}_i-y_i)^2}{\sum_i (y_i-\bar{y})^2} \\
    RMSE_{Ens} &= \sqrt{MSE_{Ens}} 
  \end{aligned}
\end{equation}
with $\bar{\hat{y}}_i$ the mean of the predicted ensemble for sample $i$, $\bar{y}$ the mean of the observations and $n$ the number of samples the score is computed over. Thus the skill scores are computed as:
\begin{equation}\label{eq:skill_scores}
\resizebox{0.9\textwidth}{!}{$
\begin{aligned}
    CRPSS &= 1 - \dfrac{CRPS_{S2S}}{CRPS_{clim}} & MSES_{Ens} &= 1 - \dfrac{MSE_{Ens,S2S}}{MSE_{Ens, clim}} & RMSES_{Ens} &= 1 - \dfrac{RMSE_{Ens,S2S}}{RMSE_{Ens, clim}}  \\
    MAES_{Ens} &= 1 - \dfrac{MAE_{Ens,S2S}}{MAE_{Ens, clim}}  &  R^{2}S_{Ens} &= \dfrac{R^2_{Ens, S2S} - R^{2}_{Ens, clim}}{1 - R^{2}_{Ens, clim}}
\end{aligned}
$}
\end{equation}
because for a perfect forecast, the CRPS, MSE, MAE, RMSE are zero while its R2 is one. The observations used here are the wind power supply for France at a daily resolution. They were obtained from the TSO open-data platform at the same time the training data for the weather-to-power model was retrieved. Only the $CRPSS$ and $MSES$ are presented in the body of the paper but other metrics can be found in appendix \ref{appendixB}.

\subsection{Post-Processing Methods}\label{sec:pp_methods}
Ensemble weather forecasts often exhibit biases and under-dispersion, leading to a poorly calibrated forecast. To compensate for these aspects, they are often post-processed using statistical post-processing methods \cite{vannitsem_statistical_2018}. However, when working with a weather-to-power forecasting pipeline, a question arises: where should the post-processing step be applied. Indeed, the weather forecasts could be post-processed prior to being turned into power forecasts, which can be post-processed as well. Four possible strategies emerge : no post-processing, post-process only the weather ensembles prior to the power conversion, post-process only the power ensembles, or post-process at both steps. Building on the work from Phipps \textit{et al.} \cite{phipps_2022} and our own experiments on the dataset, no benefit seems to appear from post-processing the weather data. Hence why, we followed a paradigm in which only the power ensemble forecast is post-processed as the last step of the pipeline.

Post-processing methods can be classified into two categories: parametric and non-parametric. Parametric models require choosing a distribution and its specification before fitting the corresponding parameters to the forecasts and observations using CRPS or likelihood optimization. Non-parametric models do not require to imply a distribution nor its parameters. Instead, the quantiles of the ensemble distribution are dressed so that they match the observations by learning directly from the data. Parametric techniques such as Ensemble Model Output Statistics (EMOS) or Bayesian Model Averaging (BMA) were applied to weather forecasts alongside the Ensemble Prediction Systems in the 2000s \cite{gneiting_calibrated_2005, raftery_using_2005}. They offer a trade-off between interpretability, efficiency, and performance. Yet, with the increase of data availability, machine learning based models showed that they were able to compete~\cite{primo_comparison_2024, schulz_machine_2022, taillardat_research_2020}.

In this work, we tried eight different post-processing methods, four parametric and four non-parametric. For the sake of clarity, only the best post-processing methods for each family are going to be explained in the body of the paper. They were chosen based on their skill (both probabilistic and deterministic) as well as their calibration, even tough some methods offers close performance. Still, the six remaining techniques are described in Appendix \ref{appendixA} and the full results can be found in Appendix \ref{appendixB} and \ref{appendixC}. For both families we created a post-processed power ensemble of the same size as the raw ensemble, i.e $51$ member, by defining $K=51$ quantiles as $q_i = \dfrac{i}{K+1}$ and either computing them via the distribution or by training the model to predict those quantiles. For parametric models, we chose a normal distribution truncated over $[0, +\infty[$ as traditionally used for parametric post-processing of power ensembles \cite{horat_improving_2025, phipps_2022}.

In order to create post-processed power ensembles for each sample of our dataset, we went with an online post-processing framework which is represented in the bottom dashed box of figure \ref{fig:flowchart}. This means that to obtain the post-processed ensemble for a given day, all the previous pairs of observation and forecasts available were used to fit the model. Since forecasts are issued up to $h$ days in advance, with $h$ being the forecast lead time, one can only use the forecasts/observations pairs issued $h$ days prior to the current forecast initialization for training the post-processing model. Doing this allows us to keep as much data as possible and mimic what would be done in an operational setting. We only miss post-processed ensembles for the first $N$ samples of the dataset. Indeed, suppose that $N$ samples are needed to fit the post-processing model, $N$ could be model-dependant, then one can not retrieve post-processed ensembles for these $N$ points. Since the data is ordered temporally, the first $N$ days won't have a post-processed ensemble but all the others starting from the $N+1$ will. In our case $N$ is used to initialize the first model, before incrementally increasing it as observations and forecasts become available. This hyperparameter can be optimized but we went with $N=30$ as several works on weather post-processing advise for a size of roughly a month, a month and a half in case of a sliding evaluation (i.e no increase of the training set size) \cite{gneiting_calibrated_2005, thorarinsdottir_probabilistic_2010}. We monitored the convergence of the methods in terms of CRPS as the training set size increases to provide recommendations to other practitioners, which are given in appendix \ref{appendixD}.

\subsubsection{Best Parametric Post-Processing Method :  Ensemble Model Output Statistics (EMOS)}\label{sec:emos}
Ensemble Model Output Statistics (EMOS) is a variant of Model Output Statistics (MOS) based on non-homogeneous regression~\cite{gneiting_calibrated_2005, vannitsem_statistical_2018}. EMOS expresses the quantity $Y$ as :
\begin{equation} \label{eq:emos_mean}
    Y = \sum_{i=1}^K a_iX_i + b + \varepsilon
\end{equation}
where $a_i$ and $b$ are regression coefficients, $X_i$ the ensemble members and $\varepsilon$ is an error term with zero mean. In case of indistinguishable members, as it is the case with ECMWF ensembles and therefore our power ensembles, the following simplification can be done: $\forall i \: a_i=a$. Here indistinguishable means that the ensemble distribution is uniform across the member i.e each member is equally probable. Given this deterministic forecast, we can create a probabilistic forecast by assuming a distribution. The mean $\mu$ and variance $\sigma^2$ of the distribution can be modeled using the mean and variance of the raw ensemble. For example, if a normal or truncated normal distribution is chosen, we use equation \ref{eq:emos_mean} to model the mean $\mu$ and approximate the variance $\sigma^2$ as a linear function of the ensemble spread $S$:
\begin{equation}
    Y_{|(X_1\ldots, X_K)} \sim \mathcal{N}(\mu = \sum_{i=1}^K a_iX_i + b, \sigma^2 = c+dS^2)
\end{equation}
with $c$ and $d$ the coefficients of the linear relationship between the distribution variance and the ensemble spread. In order to estimate the model's coefficient, one can use CRPS minimization or likelihood optimization. In case of CRPS minimization, which is often preferred, closed formulas of the CRPS for different distribution exist allowing for gradient based optimisation. The reader can refer to the appendix in Taillardat \textit{et al.} for an exhaustive list \cite{taillardat_calibrated_2016}. In our case, we optimized the CRPS of the truncated normal distribution that the post-processed ensemble $X$ should follow : $X \sim \mathcal{N}_{[0,+\infty[}(\mu, \sigma^2)$.
\begin{equation}
    CRPS(X, y) = \dfrac{\sigma}{p^2} \left\{ \omega p \left[2\Phi(\omega) + p - 2\right] + 2p\phi(\omega) - \dfrac{1}{\sqrt\pi}\Phi\left(\dfrac{\mu\sqrt{2}}{\sigma}\right)\right\}
\end{equation}
with $\omega = \dfrac{y-\mu}{\sigma}$, $p = \Phi(\mu/\sigma)$, $y$ the observation and $\phi$ and $\Phi$ are the probability and cumulative distribution function of the standard normal distribution, respectively.

The framework of EMOS can be extended to include more predictors or to depend on the lead time \cite{ casciaro2022, jobst_2024, messner_nonhomogeneous_2017}. Our experiments showed that having lead time dependant coefficients $a$, $b$, $c$ and $d$ was not improving the traditional EMOS model. So we fitted one model per lead time as it is computationally faster.

\subsubsection{Best Non-Parametric Post-Processing Method : Quantile Regression}\label{sec:qr}
Quantile Regression is a type of linear regression model that aims at modeling a specific quantile of the target distribution instead of the mean \cite{chapman_2017, kiranmoy_2019, Koenker_2005}. The equation of the model follows the same definition as a linear regression equation :
\begin{equation}
    Y = \sum_i a_iX_i+b+\varepsilon
\end{equation}
with $a_i$ and $b$ the coefficients, $X_i$ the ensemble members and $\varepsilon$ an error term with zero mean. The difference with a linear regression lies in the loss function used to estimate the coefficients. Instead of a standard mean squared error which aims at modeling the mean of the conditional distribution given the inputs $\mathbb{E}(Y|X)$, the pinball loss is used \cite{Steinwart_2011}. It is defined as :
\begin{equation}\label{eq:pinball_loss}
\mathcal{L}_q(\hat{y}, y) = \begin{cases}
        q(y-\hat{y}) & \text{for  } y\geq \hat{y}\\
        (1-q)(\hat{y}-y) & \text{for  } \hat{y}> y.
  \end{cases}
\end{equation}
with $\hat{y}$ the predicted forecast of quantile $q$ and $y$ the target. Using this loss the conditional quantile of $Y$ given $X$, $\mathbb{Q}(Y|X)(q)~=~\mathrm{inf}\left\{ y:F_{Y|X}(y) \geq q \right\}$ is modeled.  Fitting a quantile regression model for each of the different quantile of interest provides a proper dressing of the raw ensemble quantiles towards a calibrated distribution. It is really close to the Ensemble Model Output Statistics formulation but instead of defining a distribution, the focus is directly on the quantiles. It is also more versatile as different characteristics of the raw ensemble could be used in the regression equation such as the median, skewness, kurtosis or quantiles. To have a fair comparison with the parametric models we only used the raw ensemble members as the predictors.  A lead time dependency in the coefficients of the model can be introduced as well leading to a model for all the lead times instead of one model per lead time as we investigated with the Ensemble Model Output Statistics method (see \ref{sec:emos}). Our experiments showed once again that it did not provide any improvement, thus one model was fitted for each lead time available.

\section{Results \& Discussions}\label{sec:results_section}
In this section we present the results of the raw as well as of the post-processed wind power ensembles regarding the skill scores (see figures \ref{fig:CRPS_skill_best}, \ref{fig:ensMSE_skill_best}, \ref{fig:CRPS_skill_full}, \ref{fig:ensMSE_skill_full}, \ref{fig:ensRMSE_skill_full}, \ref{fig:ensMAE_skill_full}, and \ref{fig:ensR2_skill_full}) and the calibration (see figure \ref{fig:reliability_best} and \ref{fig:reliability_full}). The calibration is assessed through reliability plots which compare the theoretical quantile values to the observed probability that the observation lies below the empirical quantiles defined from the ensemble distribution.

\begin{figure}[ht!]
  \centering
  %\documentclass{standalone}
%\usepackage{tikz}
%\usetikzlibrary{patterns}
%\usepackage{pgfplots}
%\pgfplotsset{compat=1.18}
%\usepackage{xcolor}

%\definecolor{color0}{HTML}{D51317}
%\definecolor{color1}{HTML}{F39200}
%\definecolor{color2}{HTML}{EFD500}
%\definecolor{color3}{HTML}{95C11F}
%\definecolor{color4}{HTML}{007B3D}
%\definecolor{color5}{HTML}{31B7BC}
%\definecolor{color6}{HTML}{0094CD}
%\definecolor{color7}{HTML}{164194}
%\definecolor{color8}{HTML}{6F286A}

%\begin{document}

\begin{tikzpicture}
    \begin{axis}[
        width=15cm,
        height=10cm,
        xlabel={Lead time [D]},
        xtick={0,4,9,14,19,24,29,34,39,45},
        xticklabels={1, 5, 10, 15, 20, 25, 30, 35, 40, 46},
        ylabel={$CRPSS$},
        ymin=-0.5,
        ymax=0.5,
        xmin=0, xmax=45,
        grid=major,
        thick,
        mark size=2pt,
        legend columns = 1,
        legend style={at={(0.51,0.75)}, anchor=south west},
    ]   
        \addplot [
          draw=none,
          forget plot,
            %fill=gray!30,     % color and transparency
         pattern=north east lines,
        pattern color= black!40,
        ] coordinates {
        (0,-1) (6,-1) (6,1) (0,1)};

        \addplot+[ mark=none, line width=1pt, color=gray
        ]table[x=lead_time, y=s2s, col sep=comma, header=true]{Data/CRPS_skill_score.txt};
        \addlegendentry{Raw}
        
        \addplot+[ mark=none, line width=1pt, color = brown, style = solid
        ]table[x=lead_time, y=clim_bs, col sep=comma, header=true]{Data/CRPS_skill_score.txt};
        \addlegendentry{Climatological Bootstrap}

        \addplot+[ mark=none, line width=1pt, color=color0, style = solid
        ]table[x=lead_time, y=emoslt, col sep=comma, header=true]{Data/CRPS_skill_score.txt};
        \addlegendentry{Ensemble Model Output Statistics}

        \addplot+[ mark=none, line width=1pt, color=color4, style = solid
        ]table[x=lead_time, y=qrlt, col sep=comma, header=true]{Data/CRPS_skill_score.txt};
        \addlegendentry{Quantile Regression}

    \end{axis}
\end{tikzpicture}
%\end{document}
  \caption{$CRPSS$ evolution with lead time. The lead times between $1$ and $7$ days are hatched to remind that short-term weather forecasts would be more suitable to provide power forecasts at such horizons. For each lead time, the $CRPS$ is averaged over all the available samples.}
  \label{fig:CRPS_skill_best}
\end{figure}

\subsection{Skill Scores evolution with Lead Time}\label{sec:skill_scores}
In terms of skill score, $CRPSS$, $MSES_{Ens}$, $RMSES_{Ens}$ and $R^2S_{Ens}$ (see figures \ref{fig:CRPS_skill_best}, \ref{fig:ensMSE_skill_best}, \ref{fig:ensRMSE_skill_full} and \ref{fig:ensR2_skill_full}), using S2S weather forecasts instead of a climatology do not improve the skill of wind power forecasts for lead times longer than a week. Improvements can only be observed for short lead times, i.e forecast horizon shorter than \si{7} days, where short-term weather forecasts would be better suited to achieve even bigger enhancements. Indeed, for this range of lead times, dedicated short-term weather forecasts with higher resolution, different parametrisations and better accuracy at short timescales should be preferred to S2S predictions. Only $MAE$ (see figure \ref{fig:ensMAE_skill_full}) has a positive skill score after $7$ days but it rapidly becomes negative at $12$ days. This is due to the lower penalization of outliers by the metric.\\

In fact, for $CRPSS$, using S2S forecasts do not improve the climatological nor the bootstrap climatological baseline at all, no matter the lead time considered. We can see that for all the considered metrics, $CRPS$, $MSE_{Ens}$, $RMSE_{Ens}$, $MAE_{Ens}$ and $R^2_{Ens}$ the skill score is decreasing until reaching a plateau at a lead time of around $16$ days. The decrease was expected as forecast performance usually worsens as the horizon of prediction increases. But the plateau after $16$ days means that, on average, there is no change in skill of the forecasts for lead time ranging from $16$ to $46$ days.
Post-processing enables to improve the skill of the forecasts by another \si{10\percent} for both the $CRPS$ and $MSE$ based skill scores. This enhancement is again steady with the lead time increase. Both Ensemble Model Output Statistics and Quantile Regression post-processing methods show the same decrease and plateau in skill as the raw forecasts (see figures \ref{fig:CRPS_skill_best} and \ref{fig:ensMSE_skill_best}). In fact, both methods are really close in terms of skill score which is not surprising as they are also close in their formulation. One cannot determine clearly a better approach between the two. In both cases, post-processing the power forecasts allows to reach a skill score plateau of around \si{52\percent} for $CRPSS$ and \si{55\percent} for$MSES_{Ens}$ for lead times ranging from $15$ to $46$ days.\\
We used the same predictors for both models but it is possible that including other predictors extracted from the raw ensemble lead to different results. Quantile Regression would have each quantile of the post-processed ensemble depend on the additional predictors while Ensemble Model Output Statistics would only face a modification in the mean equation. 

Even tough S2S forecasts do not improve the climatological and boostrap climatological baselines, post-processing them enables to improve the skill of the forecasts by around \si{20\percent} for $CRPS$, \si{30\percent} for $MSE$,  \si{15\percent} for $RMSE$, \si{8\percent} for $MAE$ and \si{40\percent} for $R^2$ at most. This enhancement is steady as the lead time increases, even for the plateau region, and allows the post-processed ensembles to exhibit positive skill for lead times between $7$ and $16$ days. $CRPS$ is improved by \si{15\percent} to \si{5\percent}, $MSE$ by \si{20\percent} to \si{5\percent}, $RMSE$ by \si{18\percent} to \si{3\percent}, $MAE$ by \si{18\percent} to \si{5\percent} and $R^2$ by \si{40\percent} to \si{10\percent} over this range of lead times. All the post-processing methods keep showing the same decrease in skill until $16$ days with a plateau afterwards as the raw forecasts exhibited. The two best post-processing techniques, Ensemble Model Output Statistics and Quantile Regression, achieve really small improvements of the $CRPSS$ over the climatological baseline at lead time comprised between $16$ and $35$ days. For the $MSE$ no improvements for these lead times can be seen. The improvements ranges between \si{0,5\percent} at $18$ days and up to \si{3\percent} at $28$ days. Still, this skill gain is not enough to achieve skilled predictions at such horizons. Furthermore, it is the same amount of skill that the climatological bootstrap baseline provides. Only the $R^2$ based skill score show a positive skill plateau at around \si{10\percent}. The convergence of the best post-processed ensemble skills towards the climatological baselines highlights the difficulty of achieving skillfull predictions at long forecasts horizons, i.e horizons higher than two weeks, due to the chaotic nature of the atmosphere.
It is worth mentioning that more complex post-processing methods like Quantile Regression Forest or Distributional Regression Network, described in sections \ref{sec:qrf} and \ref{sec:qrn} that requires a lot of data are not able to outperform less complex techniques (see figure \ref{fig:CRPS_skill_full} and \ref{fig:ensMSE_skill_full}) on this small dataset. In fact, the dataset contains $602$ samples for each of the lead time which proved to be not enough for such models, leading to no or smaller improvements than with simpler methods. The Mixture of Experts (see section \ref{sec:moe} and figures \ref{fig:CRPS_skill_full} and \ref{fig:ensMSE_skill_full}) method is not as high performing as Ensemble Model Output Statistics or Quantile Regression and it is mainly due to the fact that since we only considered weighted average of the ensemble members as the mixture class, if the observation lies outside of the convex envelope defined by the raw ensemble, the model can not shift the post-processed ensemble to correct the bias accordingly. 

Finally, the absence of strong differences in terms of skill between both top performing methods EMOS and QR is not surprising as they are also close in their formulation. One cannot determine clearly a better approach between the two. We used the same predictors for both models but it is possible that including other predictors extracted from the raw ensemble lead to different results. Quantile Regression would have each quantile of the post-processed ensemble depend on the additional predictors while Ensemble Model Output Statistics would face a modification in the mean and variance equations.

\begin{figure}[ht!]
  \centering
  %\documentclass{standalone}
%\usepackage{tikz}
%\usetikzlibrary{patterns}
%\usepackage{pgfplots}
%\pgfplotsset{compat=1.18}
%\usepackage{xcolor}

%\definecolor{color0}{HTML}{D51317}
%\definecolor{color1}{HTML}{F39200}
%\definecolor{color2}{HTML}{EFD500}
%\definecolor{color3}{HTML}{95C11F}
%\definecolor{color4}{HTML}{007B3D}
%\definecolor{color5}{HTML}{31B7BC}
%\definecolor{color6}{HTML}{0094CD}
%\definecolor{color7}{HTML}{164194}
%\definecolor{color8}{HTML}{6F286A}

%\begin{document}

\begin{tikzpicture}
    \begin{axis}[
        width=15cm,
        height=10cm,
        xlabel={Lead time [D]},
        xtick={0,4,9,14,19,24,29,34,39,45},
        xticklabels={1, 5, 10, 15, 20, 25, 30, 35, 40, 46},
        ylabel={$MSES_{Ens}$},
        ymin=-0.5,
        ymax=0.5,
        xmin=0, xmax=45,
        grid=major,
        thick,
        mark size=2pt,
        legend columns = 1,
        legend style={at={(0.51,0.75)}, anchor=south west},
    ]   
        \addplot [
          draw=none,
          forget plot,
            %fill=gray!30,     % color and transparency
         pattern=north east lines,
        pattern color= black!40,
        ] coordinates {
        (0,-1) (6,-1) (6,1) (0,1)};

        \addplot+[ mark=none, line width=1pt, color=gray
        ]table[x=lead_time, y=s2s, col sep=comma, header=true]{Data/ensMSE_skill_score.txt};
        \addlegendentry{Raw}
        
        \addplot+[ mark=none, line width=1pt, brown = black, style = solid
        ]table[x=lead_time, y=clim_bs, col sep=comma, header=true]{Data/ensMSE_skill_score.txt};
        \addlegendentry{Climatological Bootstrap}

        \addplot+[ mark=none, line width=1pt, color=color0, style = solid
        ]table[x=lead_time, y=emoslt, col sep=comma, header=true]{Data/ensMSE_skill_score.txt};
        \addlegendentry{Ensemble Model Output Statistics}

        \addplot+[ mark=none, line width=1pt, color=color4, style = solid
        ]table[x=lead_time, y=qrlt, col sep=comma, header=true]{Data/ensMSE_skill_score.txt};
        \addlegendentry{Quantile Regression}

    \end{axis}
\end{tikzpicture}
%\end{document}
  \caption{$MSES_{Ens}$ evolution with lead time. The lead times between $1$ and $7$ days are hatched to remind that short-term weather forecasts would be more suitable to provide power forecasts at such horizons. For each lead time, the $MSE$ is computed over all the available samples.}
  \label{fig:ensMSE_skill_best}
\end{figure}

It is worth mentioning that more complex post-processing methods like Quantile Regression Forest or Distributional Regression Network, described in sections \ref{sec:qrf} and \ref{sec:qrn} that requires a lot of data are not able to outperform less complex techniques (see figure \ref{fig:CRPS_skill_full} and \ref{fig:ensMSE_skill_full}) on this small dataset. Due to the online post-processing procedure, the training set contained at most $602$ samples for each of the lead time which proved to be not enough for such models, leading to no or smaller improvements than with simpler methods. The Mixture of Experts (see section \ref{sec:moe} and figures \ref{fig:CRPS_skill_full} and \ref{fig:ensMSE_skill_full}) method is not as high performing as Ensemble Model Output Statistics or Quantile Regression and it is mainly due to the fact that since we only considered weighted average of the ensemble members as the mixture class, if the observation lies outside of the convex envelope defined by the raw ensemble, the model can not shift the post-processed ensemble to correct the bias accordingly.

\begin{figure}[ht!]
  \centering
  %\documentclass[tikz,border=2mm]{standalone}
%\usepackage{tikz}
%\usetikzlibrary{calc}
%\usepackage{pgfplots}
%\pgfplotsset{compat=1.18}
%\usepgfplotslibrary{groupplots} 
%\usepackage{xcolor}

%\definecolor{color0}{HTML}{D51317}
%\definecolor{color1}{HTML}{F39200}
%\definecolor{color2}{HTML}{EFD500}
%\definecolor{color3}{HTML}{95C11F}
%\definecolor{color4}{HTML}{007B3D}
%\definecolor{color5}{HTML}{31B7BC}
%\definecolor{color6}{HTML}{0094CD}
%\definecolor{color7}{HTML}{164194}
%\definecolor{color8}{HTML}{6F286A}

%\begin{document}

\begin{tikzpicture}
    \foreach \i/\j/\lt in {1/2/7, 2/2/10, 3/2/12, 1/1/15, 2/1/20, 3/1/25, 1/0/30, 2/0/35, 3/0/40} {
    \begin{axis}[
        legend to name=mylegendcalibrationbest,
        legend columns = 2,
        at={($(0,0) + (\i*5cm,\j*5cm)$)}, % adjust positioning
        anchor=origin,
        width=4.8cm,
        height=4.8cm,
        title={\textbf{Lead Time = \lt\,days}},
        xmin=0,
        xmax=1,
        ymin=0,
        ymax=1,
        thick,
        xlabel=Quantiles,
        ylabel=Frequency,
    ]
    \addplot+[mark=none, color=black, style = dashed, forget plot] table[x=quantile, y=quantile, col sep=comma,
       restrict expr to domain={\thisrow{lead}}{\lt:\lt}
        ]{Data/ensemble_quantile_probs.txt};

    \addplot+[mark=none, color=gray, style = solid, line width=1pt] table[x=quantile, y=s2s, col sep=comma,
    restrict expr to domain={\thisrow{lead}}{\lt:\lt}
        ]{Data/ensemble_quantile_probs.txt};
    \addlegendentry{Raw}

    \addplot+[mark=none, color=brown, style = solid, line width=1pt] table[x=quantile, y=clim_bs, col sep=comma,
    restrict expr to domain={\thisrow{lead}}{\lt:\lt}
        ]{Data/ensemble_quantile_probs.txt};
    \addlegendentry{Climatological Bootstrap}

    \addplot+[mark=none, color=color0, style = solid, line width = 1pt] table[x=quantile, y=emoslt, col sep=comma,
    restrict expr to domain={\thisrow{lead}}{\lt:\lt}
        ]{Data/ensemble_quantile_probs.txt};
    \addlegendentry{Ensemble Model Output Statistics}
    
    \addplot+[mark=none, color=color4, style = solid, line width = 1pt] table[x=quantile, y=qrlt, col sep=comma,
    restrict expr to domain={\thisrow{lead}}{\lt:\lt}
        ]{Data/ensemble_quantile_probs.txt};
    \addlegendentry{Quantile Regression}

\end{axis}
}
\end{tikzpicture}

% Place one global legend below the group
\pgfplotslegendfromname{mylegendcalibrationbest}

%\end{document}
  \caption{Reliability plots for different lead times. A perfectly calibrated forecast would align with the first bisector represented by the black dashed line. Lead times below $7$ days are not represented as short-term weather forecasts would be preferred to provide the corresponding short-term power forecasts. For each lead time, the observed frequency are computed over all the available samples. The quantile discretization step is 0,05.}
  \label{fig:reliability_best}
\end{figure}

\subsection{Calibration for different Lead Time}\label{sec:calibration}
Skillful forecasts are important but if they are unreliable they won't be trusted by their users. That's why calibration of the forecasts is crucial and should be consistent along the different horizons. Figures \ref{fig:reliability_best} and \ref{fig:reliability_full} stresses out the calibration of the forecasts for different lead times. As one can expect, the raw S2S power forecasts are not calibrated due to the biases and under-dispersion of the S2S weather forecasts that propagates to the power forecast ensembles. It is worth pointing out that the raw forecasts always under predict the quantiles for all the quantiles above $0.1$ meaning that the power forecasts are biased negatively and under predicts wind power. This could stem from weather forecasts or from the extrapolation of the \si{10 \meter} wind speed to \si{100 \meter} leading to lower predicted wind speed than in reality or it could comes from the weather-to-power model.

Despite the unreliability of the raw forecasts, applying a post-processing model dresses the quantiles accordingly leading to a calibrated forecast for all the available lead times. Both Ensemble Model Output Statistics and the Quantile Regression model offers the same level of calibration of the forecasts for most lead times leading to reliable predictions for every lead time. The quantiles below $0.4$ are slightly overestimated while the quantiles above $0.7$ are slightly underestimated. This means that the predicted quantile value is not low or high enough compared to the true corresponding observation. This could stem from the way we train the post-processing model. The wind power generation is growing steadily in France, leading to a small upward trend in the observation. By training the post-processed model using only the raw ensemble members as predictors we cannot account for this aspect.

Two techniques, Bayesian Model Averaging and Distributional Regression Network (see figure~\ref{fig:reliability_full}), failed completely to dress the quantile accordingly. They either over or under-predict the quantile values leading to an unreliable forecast. Quantile Regression Forest and Network are able to achieve reliable predictions despite the small training data and the poor improvement of the skill scores. This show that skillful and reliable forecasts are difficult to achieve jointly. Consequently, we would advise other practitioners to prefer Ensemble Model Output Statistics, Member by Member, Quantile Regression and Mixture of Experts as post-processing techniques for S2S power forecasts. Depending on the size of the data at hand, one could also consider Quantile Regression Forest or Network if the training dataset size is sufficient.

\subsection{Discussion of the Climatological Bootstrap Baseline}\label{sec:clim_bs}
We can see in figures \ref{fig:reliability_best} and \ref{fig:reliability_full} that the simple climatological bootstrap approach to improve the climatological baseline does not provide a better calibration. Small quantiles systematically have a higher frequency while higher quantiles have a lower one. When looking at the skill scores, this baseline seems to be a bit better than the standard climatological baseline. In fact, it is an artefact caused by the increased ensemble size as we demonstrate in appendix \ref{appendixE}. The climatological baseline has \si{30} members while the climatological bootstrap baseline has \si{51} members. As ensemble size matters, building a bigger and potentially better ensemble by leveraging bootstrap techniques would be recommended for future studies.

Indeed, in this work the climatological bootstrap was used as a baseline for comparison. Thus, we used a very simple bootstrap technique. Therefore, we could think of multiple sophisticated bootstrap methods to construct an ensemble of possible weather conditions for a given day out of the available past weather data. For example, instead of defining similar weather conditions using a window centered around a given date for each variable as we did, we could look at atmospheric analogues of the date in the past data. This would take into account the multivariate aspect of the weather conditions directly. We could also try to choose the analogues using the observed wind power supply for the given date. By applying such ideas, we could build weather ensembles with more members and suited toward the forecasting of wind power for a fraction of the cost of running a real S2S Ensemble Prediction System. Yet, weather forecasts at subseasonal-to-seasonal scales will continue to develop and thus improve leading to more accurate renewable power forecasts. It is also important to note that S2S weather forecasts are better at forecasting extreme weather events than a climatological proxy. This is crucial in a sustainable power system to anticipate cutoff of wind turbines during storms, dunkelflaute or hellbrise and thus ensure a stable supply.

\section{Conclusion}\label{sec:conclusion}
We showed that it was possible to leverage subseasonal-to-seasonal weather forecasts at daily resolution without any temporal nor spatial aggregation to provide daily forecasts of France wind power until $46$ days in advance. We propose a pipeline consisting of a weather-to-power model and a post-processing step. The lead time agnostic weather-to-power model transforms weather data into wind power supply directly using the location and power capacity of wind farms over France. The post-processing step allows to improve the skill as well as the reliability of the predictions by correcting the biases and dispersion of the power ensemble. 

Indeed, without post-processing the power forecasts do not offer any skill improvements over a simple climatological baseline for all the considered forecasting horizons, in addition, to being unreliable. Not only did the post-processing step enhanced the calibration of the power ensembles but it also improved the skill by \si{20\percent} for the $CRPS$ skill score, \si{30\percent} for $MSE$, \si{15\percent} for $RMSE$ and \si{40\percent} for $R^2$ for every lead time. This allowed to achieve a enhancement of \si{15\percent} to \si{5\percent} for $CRPS$ and \si{20\percent} to \si{5\percent} for $MSE$ over the lead times comprised between $7$ and $16$ days.
Still, skill at subseasonal scales is hard to obtain as we showed that a strong decrease of skill up to a lead time of $16$ days happens before reaching a plateau which is not better than the climatological baseline.

The results imply that daily renewable forecasts up to two weeks in advance are achievable, leading to progress in grid and O\&M management as well as better risk mitigation for electricity suppliers. With the expected refinement of S2S weather ensemble prediction systems, S2S power forecasts will see their skill and adoption increase contributing to a cleaner and more robust energy system.

This method is also applicable to other weather dependent power sources like solar or hydropower not to mention other weather sensitive fields such as agriculture or aviation where mid-term forecasts accounting for the weather are needed.

\noindent 
\paragraph{Acknowledgments} The authors would like to thoroughly thank Maxime Taillardat and Sebastian Lerch for the answers, tips and advice they provided during the early discussions about this project.

\paragraph{Competing Interests} The authors declare that they have no competing interests.

\paragraph{Funding} This research was supported by a grant from the Association Nationale de la Recherche et de la Technologie (ANRT) No. 2024/0010.

\clearpage

\begin{appendix}
\section{Appendix A : Post-Processing Methods}\label{appendixA}
Here the post-processing techniques that were not retained for the main discussion are developed. They are split into 2 categories: parametric and non-parametric methods. Parametric models imply specifying a probability distribution and estimating its parameters, while non-parametric models directly estimate the quantiles through proper quantile learning. In the following sections $Y$ represents the quantity of interest and $(X_1,\ldots, X_K)$ is a ensemble of $K=51$ members forecasting $Y$. For both families we created a post-processed power ensemble of the same size as the raw ensemble, i.e $51$ members, by defining $K=51$ quantiles as $q_i = \dfrac{i}{K+1}$ and either computing them via the distribution or by training the model to predict those quantiles. For parametric models, we chose a normal distribution truncated over $[0, +\infty[$ as traditionally used for parametric post-processing of power ensembles \cite{horat_improving_2025, phipps_2022}.

\subsection{Parametric Models}

\subsubsection{Bayesian Model Averaging (BMA)}\label{sec:bma}
Bayesian Model Averaging (BMA) is a technique used to combine predictions from several different models to achieve a more accurate prediction, rather than selecting the best model and potentially underestimating the uncertainty coming from the various models' predictions \cite{fragoso_bayesian_2018, hoeting_bayesian_nodate}. It is particular case of the Mixture of Experts technique (see section \ref{sec:moe}). Its first use for post-processing weather forecasts dates back to $2003$ \cite{raftery_using_2005}. Let us consider the $K$ members' predictions as different model predictions. Using the law of total probability given the training data $Y_T$:
\begin{equation}
    \mathbb{P}(Y|Y_T) = \sum_{i=1}^K\mathbb{P}(Y|X_i, Y_T)\mathbb{P}(X_i|Y_T)
\end{equation}
with $\mathbb{P}(Y|X_i)$ the forecast probability distribution based on member $X_i$ alone given that it's the best forecast and $\mathbb{P}(X_i|Y_T)$ the posterior probability distribution of model $X_i$ being correct given the training data $Y_T$. $\mathbb{P}(X_i|Y_T)$ reflects how well member $X_i$ forecast $Y$ over the training period. These posterior probabilities must add up to one, i.e, $\sum_{i=1}^K \mathbb{P}(X_i|Y_T)=1$ and can be viewed as weights. Thus, the Bayesian probability distribution is a weighted average of the conditional probability distribution given each of the individual members. The weights are the posterior probabilities which are based on the performance of member $X_i$ over the training period.\break

\noindent Usually, $\mathbb{P}(Y|X_i)$ is approximated using a chosen distribution, for example a normal distribution, with variance $\sigma_i^2$ and mean $\mu_i$ which is a linear functions of the member forecast $X_i$ : 
\begin{equation}
    Y_{|X_i} \sim \mathcal{N}(\mu_i = a_i + b_iX_i, \sigma_i^2)
\end{equation}
with $a_i$ and $b_i$ the coefficient of the linear function and $\sigma_i^2$ the variance of the distribution. This lead to the following post-processed distribution :
\begin{equation}
    Y_{|(X_1,...,X_K)}\sim \mathcal{N}\left( \mu = \sum_{i=1}^K \omega_i(a_i+b_iX_i), \sigma^2 =\sum_{i=1}^K \omega_i\left((a_i+b_iX_i)- \sum_{j=1}^K\omega_j(a_j+b_jX_j)\right)^2+\sum_{i=1}^K \omega_i\sigma_i^2  \right)
\end{equation}
where $\omega_i = \mathbb{P}(X_i|Y_T)$ are the weights. Estimating the $a_i$, $b_i$, $\omega_i$ and $\sigma_i$ is a two-step process. First the $a_i$ and $b_i$ are obtained by regressing $X_i$ on $Y$. Then $\omega_i$ and $\sigma_i$ are found trough likelihood maximization. Since the log-likelihood function cannot be maximized analytically and is hard to maximize numerically, the Expectation-Minimization algorithm is traditionally used.

\subsubsection{Member by Member (MBM)}\label{sec:mbm}
Member by Menber (MBM) offers to post-process each member individually respecting the rank correlations of the ensemble and therefore its structure \cite{demaeyer_2023, van_schaeybroeck_ensemble_2015}. To do that the following corrected member is constructed:
\begin{equation}
    \Tilde{X_i} = \alpha + \beta\bar{X} + \tau\varepsilon_i
\end{equation}
where $\alpha$ and $\beta$ are coefficients of a linear regression over the ensemble mean $\bar{X}$ and $\varepsilon_i=X_i-\bar{X}$ is the deviation of the member from the ensemble mean. $\tau$ adjust the spread of the corrected ensemble and is defined as $\tau^2 = \gamma_1^2 + \gamma_2^2S^{-2}$ with $S$ the raw ensemble spread and $\gamma_1$ the ensemble spread-scale parameter and $\gamma_2$ the ensemble-spread nudge parameter. This implies that the corrected ensemble spread is defined by : $\Tilde{\sigma}^2 = \gamma_1\sigma^2 + \gamma_2 $. Scaling refers to increasing or decreasing of the raw ensemble spread to account for under or overdispersion and nudging refers to an additive correction. Each member is post-processed individually by adjusting it to the mean and the spread of the raw ensemble. The coefficients of the model can be estimated either by likelihood optimization or using a CRPS minimization.

\subsubsection{Distributional Regression Network (DRN)}\label{sec:drn}
Distributional Regression Network (DRN) are a kind of neural networks that outputs the parameters of a chosen distribution \cite{avanzi2024, kou2019}. Using such architecture for ensemble weather forecasts post-processing was proposed by Rasp \textit{et al.} \cite{rasp_neural_2018}. The inputs which are the raw ensemble mean $\mu$ and spread $\sigma$ are fed into the model. Each layer $l$ of neurons is a linear combinations of the weights $\omega_l^i$ and biases $b_l^i$ of the previous layer inputs $z_{l-1}^i$ and can be written as $z_l^i = g\left(\sum_j \omega_l^j z_{l-1}^j + b_l^i\right)$ with $g$ a non-linear activation function. This permits the network to learn non-linear relationships between the inputs and the outputs. To post-process weather or power ensemble, one must choose a distribution and then the relationship between the raw ensemble spread and mean and the output distribution parameters, for example the mean $\Tilde{\mu}$ and variance $\Tilde{\sigma}^2$ of a normal distribution, is learned through the network. The loss function used to train the model is the CRPS of the chosen distribution which provides a proper loss for the distributional problem at hand.

\subsection{Non-Parametric Models}
\subsubsection{Mixture of Experts (MoE)}\label{sec:moe}
Aggregation of Experts or Mixture of Experts (MoE) refers to the idea that one can achieve a better forecast by combining  a set of predictions or so called experts \cite{cesabianchi_1993, freund_1997, vovk_1990}. Let $y_t$ be the target and $\hat{y}_{t,k} \: \forall k \in \left\{1,\ldots,K\right\}$ the associated forecasts or experts at time $t$. The goal is to build an aggregated forecast or mixture of experts $y^{*}_t$ using the forecasts $y_{t,k}$ as predictors to achieve a better prediction.Given a loss function $l$ that allows to measure how good the prediction is compared to the target, we can measure the performance of each expert with $l_{t,k} = l(\hat{y}_{t,k}, y_t)$, as well as the performance of the mixture $l^{*} = l(y^{*}_t, y_t)$. Since the goal is to combine the experts to achieve a better prediction, a natural way to do so is to compare the quality of the mixture $y^{*}_t$ with the best expert available. The best expert would be defined as the one having the best performance over a fixed period of past time steps $\left[1,T\right]$ with $T\leq t-1$ for which $(y_1,\ldots, y_{T})$ and the corresponding forecasts $y_{i,k} \: \forall i \in [1,T]$ are available.\break

\noindent To define the best performance over the period $T$, one can look at the cumulative loss $L_{[1,T]}(k) = \sum_{i=1}^{T} l_{i,k}$. Thus the best expert is the one that minimizes $L_{[1,T]}$ and we can introduce the regret $R_{[1,T]}$ defined as :
\begin{equation}\label{eq:regret}
    R_{[1,T]} = L_{[1,T]}^{*} - \min_{k\in\{1,\ldots,K\}}L_{[1,T]}(k)
\end{equation}
where $L_{[1,T]}^{*}$ is the cumulative loss of the mixture. It represents the regret of "not having listened enough" to the best expert for building the mixture. The lower the regret, the better the mixture and if the regret is negative, it means that the mixture of experts outperforms the best expert and therefore any expert. Most of the classical mixture procedure involves optimizing the regret or an upper bound to obtain the mixture prediction.\break

\noindent Several classes of procedure can be considered to build to mixture but we only considered the weighted average class which defines $y^{*}_t$ as :
\begin{equation}
    y^{*}_t = \sum_{k=1}^{K} \omega_{t,k}\hat{y}_{t,k}
\end{equation}
with $\omega_{t,k}$ the weight given to expert $k$ at time $t$ that must adds up to one, i.e $\sum_{k=1}^K \omega_{t,k} = 1$. Deriving the weights can be done following different algorithms that can be static or dynamic. Static means that the set of weights learned on the past data is fixed, i.e $\omega_{t,k} = \omega_{k}$ while dynamic implies a regular update of the weights as in online learning techniques. In our case of power forecasts post-processing, the experts are the ensemble members and the loss function is the pinball loss (see equation \ref{eq:pinball_loss}) for each quantile of the post-processed ensemble. Using this loss, the conditional quantile of $Y$ given $X$, $\mathbb{Q}(Y|X)(q)~=~\mathrm{inf}\left\{ y:F_{Y|X}(y) \geq q \right\}$ is modeled. That said, for each quantile $q$ of the post-processed ensemble and each time step $t$ we build a mixture of the raw ensemble members $y^{*}_t(q)$ that best predicts this quantile. This implies that one mixture model is trained for each quantile. This procedure is repeated for each lead time. We used the opera package built by Gaillard \textit{et al.} which implements several weights estimation algorithm suited with the online learning framework \cite{gaillard2016opera}. 

\subsubsection{Quantile Regression Forest (QRF)}\label{sec:qrf}
Quantile Regression Forest (QRF) is an extension of the well known Random Forest approach \cite{meinshausen_quantile_2006, taillardat_calibrated_2016}. Instead of predicting the conditional mean of the target $Y$ given the inputs $X$, $\mathbb{E}(Y|X)$, quantile forest predicts specific quantiles $q$ of the target's distribution. The difference lies in the way the decision trees are built. For each final leaf of the trees, one does not compute the mean of the target's value falling inside this leaf but rather compute the entire empirical cumulative distribution function. Then, when doing inference, after following the binary splitting to find the associated leaf in each tree the final forecast is obtained  by averaging all the CDFs from all the trees. By construction, the final CDF is bounded by the learning sample range. Finally, the forecasted set of quantiles is obtained thanks to the predicted CDF of the forest. This model requires a lot more data to learn the relationship between the raw ensemble mean and variance and the post-processed ensemble quantiles. So we trained one model for all the lead times available by using the lead time as an input feature. Other features based on the characteristics of the raw ensemble could be used in the such as the median, skewness, kurtosis or quantiles. But to have a fair comparison with the parametric models we only used the raw ensemble members as the predictors

\subsubsection{Quantile Regression Network (QRN)}\label{sec:qrn}
Quantile Regression Network (QRN) is a neural network that transforms the inputs into a forecast of a set of quantiles \cite{cannon_2011, taylor_quantile_2000}. The output is the set of desired quantiles, and the network is trained using a generalization of the pinball loss (\ref{eq:pinball_loss}). Each predicted quantile is evaluated with regards to the target using the pinball loss for a given quantile $q$. The final loss is obtained by averaging all the pinball losses. QRNs are flexible and efficient at modeling quantiles, but they do not ensure non-crossing quantiles \cite{decke2024efficientmultiquantileregression}. Several techniques were discussed to force the predicted quantiles to be ordered. We choose to predict the increments in between the quantiles before reconstructing them using a cumulative sum. This last step ensured ordered quantiles. As for the Quantile Regression Forest (see section \ref{sec:qrf}) we build one model for all the lead times by including it as an input feature of the model and we only compared the raw ensemble members as features.

\clearpage
\section{Appendix B : Skill Scores evolution with Lead Time for the eight tested post-processing methods}\label{appendixB}
\begin{figure}[ht!]
  \centering
  %\documentclass[tikz,border=2mm]{standalone}
%\usepackage{tikz}
%\usetikzlibrary{patterns}
%\usepackage{pgfplots}
%\pgfplotsset{compat=1.18}
%\usepackage{xcolor}

%\definecolor{color0}{HTML}{D51317}
%\definecolor{color1}{HTML}{F39200}
%\definecolor{color2}{HTML}{EFD500}
%\definecolor{color3}{HTML}{95C11F}
%\definecolor{color4}{HTML}{007B3D}
%\definecolor{color5}{HTML}{31B7BC}
%\definecolor{color6}{HTML}{0094CD}
%\definecolor{color7}{HTML}{164194}
%\definecolor{color8}{HTML}{6F286A}
%\begin{document}

\begin{tikzpicture}
    \begin{axis}[
        width=15cm,
        height=10cm,
        xlabel={Lead time [D]},
        xtick={0,4,9,14,19,24,29,34,39,45},
        xticklabels={1, 5, 10, 15, 20, 25, 30, 35, 40, 46},
        ylabel={$CRPSS$},
        ymin=-0.5,
        ymax=0.5,
        xmin=0, xmax=45,
        grid=major,
        thick,
        mark size=2pt,
        legend columns = 2,
        legend style={at={(0.03, -0.4)}, anchor=south west},
    ]   
        \addplot [
          draw=none,
          forget plot,
            %fill=gray!30,     % color and transparency
         pattern=north east lines,
        pattern color= black!40,
        ] coordinates {
        (0,-1) (6,-1) (6,1) (0,1)};

        \addplot+[ mark=none, line width=1pt, color=gray
        ]table[x=lead_time, y=s2s, col sep=comma, header=true]{Data/CRPS_skill_score.txt};
        \addlegendentry{Raw}
        
        \addplot+[ mark=none, line width=1pt, color = brown, style = solid
        ]table[x=lead_time, y=clim_bs, col sep=comma, header=true]{Data/CRPS_skill_score.txt};
        \addlegendentry{Climatological Bootstrap}

        \addplot+[ mark=none, line width=1pt, color=color0, style = solid
        ]table[x=lead_time, y=emoslt, col sep=comma, header=true]{Data/CRPS_skill_score.txt};
        \addlegendentry{Ensemble Model Output Statistics}

        \addplot+[ mark=none, line width=1pt, color=color1, style = solid
        ]table[x=lead_time, y=bma, col sep=comma, header=true]{Data/CRPS_skill_score.txt};
        \addlegendentry{Bayesian Model Average}

        \addplot+[ mark=none, line width=1pt, color=color2, style = solid
        ]table[x=lead_time, y=mbm, col sep=comma, header=true]{Data/CRPS_skill_score.txt};
        \addlegendentry{Member By Member}
        
        \addplot+[ mark=none, line width=1pt, color=color3, style = solid
        ]table[x=lead_time, y=drn, col sep=comma, header=true]{Data/CRPS_skill_score.txt};
        \addlegendentry{Distributional Regression Network}

        \addplot+[ mark=none, line width=1pt, color=color4, style = solid
        ]table[x=lead_time, y=qrlt, col sep=comma, header=true]{Data/CRPS_skill_score.txt};
        \addlegendentry{Quantile Regression}
        
        \addplot+[ mark=none, line width=1pt, color=color5, style = solid
        ]table[x=lead_time, y=moe, col sep=comma, header=true]{Data/CRPS_skill_score.txt};
        \addlegendentry{Mixture of Experts}

        \addplot+[ mark=none, line width=1pt, color=color6, style = solid
        ]table[x=lead_time, y=qrf, col sep=comma, header=true]{Data/CRPS_skill_score.txt};
        \addlegendentry{Quantile Regression Forest}
    
        \addplot+[ mark=none, line width=1pt, color=color7, style = solid
        ]table[x=lead_time, y=qrn, col sep=comma, header=true]{Data/CRPS_skill_score.txt};
        \addlegendentry{Quantile Regression Network}

    \end{axis}
\end{tikzpicture}
%\end{document}
  \caption{$CRPSS$ evolution with lead time. The lead times between $1$ and $7$ days are hatched to remind that short-term weather forecasts would be more suitable to provide power forecasts at such horizons. For each lead time, the $CRPS$ is computed over all the available samples.}
  \label{fig:CRPS_skill_full}
\end{figure}

\clearpage
\begin{figure}[ht!]
  \centering
  %\documentclass[tikz,border=2mm]{standalone}
%\usepackage{tikz}
%\usetikzlibrary{patterns}
%\usepackage{pgfplots}
%\pgfplotsset{compat=1.18}
%\usepackage{xcolor}

%\definecolor{color0}{HTML}{D51317}
%\definecolor{color1}{HTML}{F39200}
%\definecolor{color2}{HTML}{EFD500}
%\definecolor{color3}{HTML}{95C11F}
%\definecolor{color4}{HTML}{007B3D}
%\definecolor{color5}{HTML}{31B7BC}
%\definecolor{color6}{HTML}{0094CD}
%\definecolor{color7}{HTML}{164194}
%\definecolor{color8}{HTML}{6F286A}
%\begin{document}

\begin{tikzpicture}
    \begin{axis}[
        width=15cm,
        height=10cm,
        xlabel={Lead time [D]},
        xtick={0,4,9,14,19,24,29,34,39,45},
        xticklabels={1, 5, 10, 15, 20, 25, 30, 35, 40, 46},
        ylabel={$MSES_{Ens}$},
        ymin=-0.6,
        ymax=0.6,
        xmin=0, xmax=45,
        grid=major,
        thick,
        mark size=2pt,
        legend columns = 2,
        legend style={at={(0.03,-0.4)}, anchor=south west},
    ]   
        \addplot [
          draw=none,
          forget plot,
            %fill=gray!30,     % color and transparency
         pattern=north east lines,
        pattern color= black!40,
        ] coordinates {
        (0,-1) (6,-1) (6,1) (0,1)};

        \addplot+[ mark=none, line width=1pt, color=gray
        ]table[x=lead_time, y=s2s, col sep=comma, header=true]{Data/ensMSE_skill_score.txt};
        \addlegendentry{Raw}
        
        \addplot+[ mark=none, line width=1pt, color = brown, style = solid
        ]table[x=lead_time, y=clim_bs, col sep=comma, header=true]{Data/ensMSE_skill_score.txt};
        \addlegendentry{Climatological Bootstrap}

        \addplot+[ mark=none, line width=1pt, color=color0, style = solid
        ]table[x=lead_time, y=emoslt, col sep=comma, header=true]{Data/ensMSE_skill_score.txt};
        \addlegendentry{Ensemble Model Output Statistics}

        \addplot+[ mark=none, line width=1pt, color=color1, style = solid
        ]table[x=lead_time, y=bma, col sep=comma, header=true]{Data/ensMSE_skill_score.txt};
        \addlegendentry{Bayesian Model Average}

        \addplot+[ mark=none, line width=1pt, color=color2, style = solid
        ]table[x=lead_time, y=mbm, col sep=comma, header=true]{Data/ensMSE_skill_score.txt};
        \addlegendentry{Member By Member}
        
        \addplot+[ mark=none, line width=1pt, color=color3, style = solid
        ]table[x=lead_time, y=drn, col sep=comma, header=true]{Data/ensMSE_skill_score.txt};
        \addlegendentry{Distributional Regression Network}
        
        \addplot+[ mark=none, line width=1pt, color=color4, style = solid
        ]table[x=lead_time, y=qrlt, col sep=comma, header=true]{Data/ensMSE_skill_score.txt};
        \addlegendentry{Quantile Regression}
        
        \addplot+[ mark=none, line width=1pt, color=color5, style = solid
        ]table[x=lead_time, y=moe, col sep=comma, header=true]{Data/ensMSE_skill_score.txt};
        \addlegendentry{Mixture of Experts}

        \addplot+[ mark=none, line width=1pt, color=color6, style = solid
        ]table[x=lead_time, y=qrf, col sep=comma, header=true]{Data/ensMSE_skill_score.txt};
        \addlegendentry{Quantile Regression Forest}
        
        \addplot+[ mark=none, line width=1pt, color=color7, style = solid
        ]table[x=lead_time, y=qrn, col sep=comma, header=true]{Data/ensMSE_skill_score.txt};
        \addlegendentry{Quantile Regression Network}

    \end{axis}
\end{tikzpicture}
%\end{document}
  \caption{$MSES_{Ens}$ evolution with lead time. The lead times between $1$ and $7$ days are hatched to remind that short-term weather forecasts would be more suitable to provide power forecasts at such horizons. For each lead time, the $MSE$ is computed over all the available samples.}
  \label{fig:ensMSE_skill_full}
\end{figure}

\clearpage
\begin{figure}[ht!]
  \centering
  %\documentclass[tikz,border=2mm]{standalone}
%\usepackage{tikz}
%\usetikzlibrary{patterns}
%\usepackage{pgfplots}
%\pgfplotsset{compat=1.18}
%\usepackage{xcolor}

%\definecolor{color0}{HTML}{D51317}
%\definecolor{color1}{HTML}{F39200}
%\definecolor{color2}{HTML}{EFD500}
%\definecolor{color3}{HTML}{95C11F}
%\definecolor{color4}{HTML}{007B3D}
%\definecolor{color5}{HTML}{31B7BC}
%\definecolor{color6}{HTML}{0094CD}
%\definecolor{color7}{HTML}{164194}
%\definecolor{color8}{HTML}{6F286A}
%\begin{document}

\begin{tikzpicture}
    \begin{axis}[
        width=15cm,
        height=10cm,
        xlabel={Lead time [D]},
        xtick={0,4,9,14,19,24,29,34,39,45},
        xticklabels={1, 5, 10, 15, 20, 25, 30, 35, 40, 46},
        ylabel={$RMSES_{Ens}$},
        ymin=-0.5,
        ymax=0.5,
        xmin=0, xmax=45,
        grid=major,
        thick,
        mark size=2pt,
        legend columns = 2,
        legend style={at={(0.03,-0.4)}, anchor=south west},
    ]   
        \addplot [
          draw=none,
          forget plot,
            %fill=gray!30,     % color and transparency
         pattern=north east lines,
        pattern color= black!40,
        ] coordinates {
        (0,-1) (6,-1) (6,1) (0,1)};

        \addplot+[ mark=none, line width=1pt, color=gray
        ]table[x=lead_time, y=s2s, col sep=comma, header=true]{Data/ensRMSE_skill_score.txt};
        \addlegendentry{Raw}
        
        \addplot+[ mark=none, line width=1pt, color = brown, style = solid
        ]table[x=lead_time, y=clim_bs, col sep=comma, header=true]{Data/ensRMSE_skill_score.txt};
        \addlegendentry{Climatological Bootstrap}

        \addplot+[ mark=none, line width=1pt, color=color0, style = solid
        ]table[x=lead_time, y=emoslt, col sep=comma, header=true]{Data/ensRMSE_skill_score.txt};
        \addlegendentry{Ensemble Model Output Statistics}

        \addplot+[ mark=none, line width=1pt, color=color1, style = solid
        ]table[x=lead_time, y=bma, col sep=comma, header=true]{Data/ensRMSE_skill_score.txt};
        \addlegendentry{Bayesian Model Average}

        \addplot+[ mark=none, line width=1pt, color=color2, style = solid
        ]table[x=lead_time, y=mbm, col sep=comma, header=true]{Data/ensRMSE_skill_score.txt};
        \addlegendentry{Member By Member}
        
        \addplot+[ mark=none, line width=1pt, color=color3, style = solid
        ]table[x=lead_time, y=drn, col sep=comma, header=true]{Data/ensRMSE_skill_score.txt};
        \addlegendentry{Distributional Regression Network}
        
        \addplot+[ mark=none, line width=1pt, color=color4, style = solid
        ]table[x=lead_time, y=qrlt, col sep=comma, header=true]{Data/ensRMSE_skill_score.txt};
        \addlegendentry{Quantile Regression}
        
        \addplot+[ mark=none, line width=1pt, color=color5, style = solid
        ]table[x=lead_time, y=moe, col sep=comma, header=true]{Data/ensRMSE_skill_score.txt};
        \addlegendentry{Mixture of Experts}

        \addplot+[ mark=none, line width=1pt, color=color6, style = solid
        ]table[x=lead_time, y=qrf, col sep=comma, header=true]{Data/ensRMSE_skill_score.txt};
        \addlegendentry{Quantile Regression Forest}
        
        \addplot+[ mark=none, line width=1pt, color=color7, style = solid
        ]table[x=lead_time, y=qrn, col sep=comma, header=true]{Data/ensRMSE_skill_score.txt};
        \addlegendentry{Quantile Regression Network}

    \end{axis}
\end{tikzpicture}
%\end{document}
  \caption{$RMSES_{Ens}$ evolution with lead time. The lead times between $1$ and $7$ days are hatched to remind that short-term weather forecasts would be more suitable to provide power forecasts at such horizons. For each lead time, the $RMSE$ is computed over all the available samples.}
  \label{fig:ensRMSE_skill_full}
\end{figure}

\clearpage
\begin{figure}[ht!]
  \centering
  %\documentclass[tikz,border=2mm]{standalone}
%\usepackage{tikz}
%\usetikzlibrary{patterns}
%\usepackage{pgfplots}
%\pgfplotsset{compat=1.18}
%\usepackage{xcolor}

%\definecolor{color0}{HTML}{D51317}
%\definecolor{color1}{HTML}{F39200}
%\definecolor{color2}{HTML}{EFD500}
%\definecolor{color3}{HTML}{95C11F}
%\definecolor{color4}{HTML}{007B3D}
%\definecolor{color5}{HTML}{31B7BC}
%\definecolor{color6}{HTML}{0094CD}
%\definecolor{color7}{HTML}{164194}
%\definecolor{color8}{HTML}{6F286A}
%\begin{document}

\begin{tikzpicture}
    \begin{axis}[
        width=15cm,
        height=10cm,
        xlabel={Lead time [D]},
        xtick={0,4,9,14,19,24,29,34,39,45},
        xticklabels={1, 5, 10, 15, 20, 25, 30, 35, 40, 46},
        ylabel={$MAES_{Ens}$},
        ymin=-0.4,
        ymax=0.4,
        xmin=0, xmax=45,
        grid=major,
        thick,
        mark size=2pt,
        legend columns = 2,
        legend style={at={(0.03,-0.4)}, anchor=south west},
    ]   
        \addplot [
          draw=none,
          forget plot,
            %fill=gray!30,     % color and transparency
         pattern=north east lines,
        pattern color= black!40,
        ] coordinates {
        (0,-1) (6,-1) (6,1) (0,1)};

        \addplot+[ mark=none, line width=1pt, color=gray
        ]table[x=lead_time, y=s2s, col sep=comma, header=true]{Data/ensMAE_skill_score.txt};
        \addlegendentry{Raw}
        
        \addplot+[ mark=none, line width=1pt, color = brown, style = solid
        ]table[x=lead_time, y=clim_bs, col sep=comma, header=true]{Data/ensMAE_skill_score.txt};
        \addlegendentry{Climatological Bootstrap}

        \addplot+[ mark=none, line width=1pt, color=color0, style = solid
        ]table[x=lead_time, y=emoslt, col sep=comma, header=true]{Data/ensMAE_skill_score.txt};
        \addlegendentry{Ensemble Model Output Statistics}

        \addplot+[ mark=none, line width=1pt, color=color1, style = solid
        ]table[x=lead_time, y=bma, col sep=comma, header=true]{Data/ensMAE_skill_score.txt};
        \addlegendentry{Bayesian Model Average}

        \addplot+[ mark=none, line width=1pt, color=color2, style = solid
        ]table[x=lead_time, y=mbm, col sep=comma, header=true]{Data/ensMAE_skill_score.txt};
        \addlegendentry{Member By Member}
        
        \addplot+[ mark=none, line width=1pt, color=color3, style = solid
        ]table[x=lead_time, y=drn, col sep=comma, header=true]{Data/ensMAE_skill_score.txt};
        \addlegendentry{Distributional Regression Network}
        
        \addplot+[ mark=none, line width=1pt, color=color4, style = solid
        ]table[x=lead_time, y=qrlt, col sep=comma, header=true]{Data/ensMAE_skill_score.txt};
        \addlegendentry{Quantile Regression}
        
        \addplot+[ mark=none, line width=1pt, color=color5, style = solid
        ]table[x=lead_time, y=moe, col sep=comma, header=true]{Data/ensMAE_skill_score.txt};
        \addlegendentry{Mixture of Experts}

        \addplot+[ mark=none, line width=1pt, color=color6, style = solid
        ]table[x=lead_time, y=qrf, col sep=comma, header=true]{Data/ensMAE_skill_score.txt};
        \addlegendentry{Quantile Regression Forest}
        
        \addplot+[ mark=none, line width=1pt, color=color7, style = solid
        ]table[x=lead_time, y=qrn, col sep=comma, header=true]{Data/ensMAE_skill_score.txt};
        \addlegendentry{Quantile Regression Network}

    \end{axis}
\end{tikzpicture}
%\end{document}
  \caption{$MAES_{Ens}$ evolution with lead time. The lead times between $1$ and $7$ days are hatched to remind that short-term weather forecasts would be more suitable to provide power forecasts at such horizons. For each lead time, the $MAE$ is computed over all the available samples.}
  \label{fig:ensMAE_skill_full}
\end{figure}

\clearpage
\begin{figure}[ht!]
  \centering
  %\documentclass[tikz,border=2mm]{standalone}
%\usepackage{tikz}
%\usetikzlibrary{patterns}
%\usepackage{pgfplots}
%\pgfplotsset{compat=1.18}
%\usepackage{xcolor}

%\definecolor{color0}{HTML}{D51317}
%\definecolor{color1}{HTML}{F39200}
%\definecolor{color2}{HTML}{EFD500}
%\definecolor{color3}{HTML}{95C11F}
%\definecolor{color4}{HTML}{007B3D}
%\definecolor{color5}{HTML}{31B7BC}
%\definecolor{color6}{HTML}{0094CD}
%\definecolor{color7}{HTML}{164194}
%\definecolor{color8}{HTML}{6F286A}
%\begin{document}

\begin{tikzpicture}
    \begin{axis}[
        width=15cm,
        height=10cm,
        xlabel={Lead time [D]},
        xtick={0,4,9,14,19,24,29,34,39,45},
        xticklabels={1, 5, 10, 15, 20, 25, 30, 35, 40, 46},
        ylabel={$R^2S_{Ens}$},
        ymin=-0.6,
        ymax=0.6,
        xmin=0, xmax=45,
        grid=major,
        thick,
        mark size=2pt,
        legend columns = 2,
        legend style={at={(0.03,-0.4)}, anchor=south west},
    ]   
        \addplot [
          draw=none,
          forget plot,
            %fill=gray!30,     % color and transparency
         pattern=north east lines,
        pattern color= black!40,
        ] coordinates {
        (0,-1) (6,-1) (6,1) (0,1)};

        \addplot+[ mark=none, line width=1pt, color=gray
        ]table[x=lead_time, y=s2s, col sep=comma, header=true]{Data/ensR2_skill_score.txt};
        \addlegendentry{Raw}
        
        \addplot+[ mark=none, line width=1pt, color = brown, style = solid
        ]table[x=lead_time, y=clim_bs, col sep=comma, header=true]{Data/ensR2_skill_score.txt};
        \addlegendentry{Climatological Bootstrap}

        \addplot+[ mark=none, line width=1pt, color=color0, style = solid
        ]table[x=lead_time, y=emoslt, col sep=comma, header=true]{Data/ensR2_skill_score.txt};
        \addlegendentry{Ensemble Model Output Statistics}

        \addplot+[ mark=none, line width=1pt, color=color1, style = solid
        ]table[x=lead_time, y=bma, col sep=comma, header=true]{Data/ensR2_skill_score.txt};
        \addlegendentry{Bayesian Model Average}

        \addplot+[ mark=none, line width=1pt, color=color2, style = solid
        ]table[x=lead_time, y=mbm, col sep=comma, header=true]{Data/ensR2_skill_score.txt};
        \addlegendentry{Member By Member}
        
        \addplot+[ mark=none, line width=1pt, color=color3, style = solid
        ]table[x=lead_time, y=drn, col sep=comma, header=true]{Data/ensR2_skill_score.txt};
        \addlegendentry{Distributional Regression Network}
        
        \addplot+[ mark=none, line width=1pt, color=color4, style = solid
        ]table[x=lead_time, y=qrlt, col sep=comma, header=true]{Data/ensR2_skill_score.txt};
        \addlegendentry{Quantile Regression}
        
        \addplot+[ mark=none, line width=1pt, color=color5, style = solid
        ]table[x=lead_time, y=moe, col sep=comma, header=true]{Data/ensR2_skill_score.txt};
        \addlegendentry{Mixture of Experts}

        \addplot+[ mark=none, line width=1pt, color=color6, style = solid
        ]table[x=lead_time, y=qrf, col sep=comma, header=true]{Data/ensR2_skill_score.txt};
        \addlegendentry{Quantile Regression Forest}
        
        \addplot+[ mark=none, line width=1pt, color=color7, style = solid
        ]table[x=lead_time, y=qrn, col sep=comma, header=true]{Data/ensR2_skill_score.txt};
        \addlegendentry{Quantile Regression Network}

    \end{axis}
\end{tikzpicture}
%\end{document}
  \caption{$R^2S_{Ens}$ evolution with lead time. The lead times between $1$ and $7$ days are hatched to remind that short-term weather forecasts would be more suitable to provide power forecasts at such horizons. For each lead time, the $R^2$ is computed over all the available samples.}
  \label{fig:ensR2_skill_full}
\end{figure}

\clearpage
\newgeometry{top=2cm,bottom=2cm,textheight=9in,textwidth=18cm}
\section{Appendix C : Reliability plots for the eight tested post-processing methods}\label{appendixC}
\begin{figure}[ht!]
  \centering
  %\documentclass[tikz,border=2mm]{standalone}
%\usepackage{tikz}
%\usetikzlibrary{patterns}
%\usetikzlibrary{calc}
%\usepackage{pgfplots}
%\pgfplotsset{compat=1.18}
%\usepgfplotslibrary{groupplots} 
%\usepackage{xcolor}
%\usepackage{subcaption}

%\definecolor{color0}{HTML}{D51317}
%\definecolor{color1}{HTML}{F39200}
%\definecolor{color2}{HTML}{EFD500}
%\definecolor{color3}{HTML}{95C11F}
%\definecolor{color4}{HTML}{007B3D}
%\definecolor{color5}{HTML}{31B7BC}
%\definecolor{color6}{HTML}{0094CD}
%\definecolor{color7}{HTML}{164194}
%\definecolor{color8}{HTML}{6F286A}

%\begin{document}

\begin{tikzpicture}
    \foreach \i/\j/\lt in {1/2/7, 2/2/10, 3/2/12, 1/1/15, 2/1/20, 3/1/25, 1/0/30, 2/0/35, 3/0/40} {
    \begin{axis}[
        legend to name=mylegendcalibrationfull,
        legend columns = 2,
        legend style={at={(0.03, -0.4)}, anchor=south west},
        at={($(0,0) + (\i*5cm,\j*5cm)$)}, % adjust positioning
        anchor=origin,
        width=4.8cm,
        height=4.8cm,
        title={\textbf{Lead Time = \lt\,days}},
        xmin=0,
        xmax=1,
        ymin=0,
        ymax=1,
        thick,
        xlabel=Quantiles,
        ylabel=Frequency,
    ]
    \addplot+[mark=none, color=black, style = dashed, forget plot] table[x=quantile, y=quantile, col sep=comma,
       restrict expr to domain={\thisrow{lead}}{\lt:\lt}
        ]{Data/ensemble_quantile_probs.txt};

    \addplot+[mark=none, color=gray, style = solid, line width=1pt] table[x=quantile, y=s2s, col sep=comma,
    restrict expr to domain={\thisrow{lead}}{\lt:\lt}
        ]{Data/ensemble_quantile_probs.txt};
    \addlegendentry{Raw}
    
    \addplot+[mark=none, color=brown, style = solid, line width=1pt] table[x=quantile, y=s2s, col sep=comma,
    restrict expr to domain={\thisrow{lead}}{\lt:\lt}
        ]{Data/ensemble_quantile_probs.txt};
    \addlegendentry{Climatological Bootstrap}

    \addplot+[mark=none, color=color0, style = solid, line width = 1pt] table[x=quantile, y=emoslt, col sep=comma,
    restrict expr to domain={\thisrow{lead}}{\lt:\lt}
        ]{Data/ensemble_quantile_probs.txt};
    \addlegendentry{Ensemble Model Output Statistics}
    
    \addplot+[mark=none, color=color1, style = solid, line width = 1pt] table[x=quantile, y=bma, col sep=comma,
    restrict expr to domain={\thisrow{lead}}{\lt:\lt}
        ]{Data/ensemble_quantile_probs.txt};
    \addlegendentry{Bayesian Model Averaging}
    
    \addplot+[mark=none, color=color2, style = solid, line width = 1pt] table[x=quantile, y=mbm, col sep=comma,
    restrict expr to domain={\thisrow{lead}}{\lt:\lt}
        ]{Data/ensemble_quantile_probs.txt};
    \addlegendentry{Member by Member}
    
    \addplot+[mark=none, color=color3, style = solid, line width = 1pt] table[x=quantile, y=drn, col sep=comma,
    restrict expr to domain={\thisrow{lead}}{\lt:\lt}
        ]{Data/ensemble_quantile_probs.txt};
    \addlegendentry{Distributional Regression Network}
    
    \addplot+[mark=none, color=color4, style = solid, line width = 1pt] table[x=quantile, y=qrlt, col sep=comma,
    restrict expr to domain={\thisrow{lead}}{\lt:\lt}
        ]{Data/ensemble_quantile_probs.txt};
    \addlegendentry{Quantile Regression }
    
    \addplot+[mark=none, color=color5, style = solid, line width = 1pt] table[x=quantile, y=moe, col sep=comma,
    restrict expr to domain={\thisrow{lead}}{\lt:\lt}
        ]{Data/ensemble_quantile_probs.txt};
    \addlegendentry{Mixture of Experts}

    \addplot+[mark=none, color=color6, style = solid, line width = 1pt] table[x=quantile, y=qrf, col sep=comma,
    restrict expr to domain={\thisrow{lead}}{\lt:\lt}
        ]{Data/ensemble_quantile_probs.txt};
    \addlegendentry{Quantile Regression Forest}
    
    \addplot+[mark=none, color=color7, style = solid, line width = 1pt] table[x=quantile, y=qrn, col sep=comma,
    restrict expr to domain={\thisrow{lead}}{\lt:\lt}
        ]{Data/ensemble_quantile_probs.txt};
    \addlegendentry{Quantile Regression Network}

\end{axis}
}
\end{tikzpicture}
% Place one global legend below the group
\pgfplotslegendfromname{mylegendcalibrationfull}

%\end{document}
  \caption{Reliability plots for different lead times. A perfectly calibrated forecast would align with the first bisector represented by the black dashed line. Lead times below $7$ days are not represented as short-term weather forecasts would be preferred to provide the corresponding short-term power forecasts. For each lead time, the observed frequency are computed over all the available samples.The quantile discretization step is 0,05.}
  \label{fig:reliability_full}
\end{figure}
\restoregeometry

\clearpage
 \newgeometry{
    textheight=9in,
    textwidth=18cm,
    top=1in,
    headheight=24pt,
    headsep=25pt,
    footskip=30pt,
    bottom = 1in,
  }
\section{Appendix D : Convergence of the post-processing methods as the training set size increases}\label{appendixD}
\begin{figure}[ht!]
  \centering
  %\documentclass[tikz,border=2mm]{standalone}
%\usepackage{tikz}
%\usetikzlibrary{patterns}
%\usetikzlibrary{calc}
%\usepackage{pgfplots}
%\pgfplotsset{compat=1.18}
%\usepgfplotslibrary{groupplots} 
%\usepackage{xcolor}
%\usepackage{subcaption}

%\definecolor{color0}{HTML}{D51317}
%\definecolor{color1}{HTML}{F39200}
%\definecolor{color2}{HTML}{EFD500}
%\definecolor{color3}{HTML}{95C11F}
%\definecolor{color4}{HTML}{007B3D}
%\definecolor{color5}{HTML}{31B7BC}
%\definecolor{color6}{HTML}{0094CD}
%\definecolor{color7}{HTML}{164194}
%\definecolor{color8}{HTML}{6F286A}

%\begin{document}

\begin{tikzpicture}
    \begin{axis}[
        legend to name=mylegendconv,
        legend columns = 2,
        width=15cm,
        height=10cm,
        grid=major,
        thick,
        mark size=2pt,
        xlabel={Training Set Size [D]},
        ylabel={$CRPS$},
        xmin=30,
        xmax=555,
        ymin = 800,
        ymax = 4500,
        xtick={30,100,150,200,250,300,350,400,450,500,555},
        legend columns = 2,
        legend style={at={(0.03, -0.4)}, anchor=south west},
        ]
    \addplot+[mark=none, color=color0, style = solid, opacity = 0.75] table[x=training_window_length, y=crps_emoslt, col sep=comma
        ]{Data/crps_training_window_length.txt};
    \addlegendentry{Ensemble Model Output Statistics}
    
    \addplot+[mark=none, color=color1, style = solid, opacity = 0.75] table[x=training_window_length, y=crps_bma, col sep=comma
        ]{Data/crps_training_window_length.txt};
    \addlegendentry{Bayesian Model Averaging}

    \addplot+[mark=none, color=color2, style = solid, opacity = 0.75] table[x=training_window_length, y=crps_mbm, col sep=comma
        ]{Data/crps_training_window_length.txt};
    \addlegendentry{Member by Member}

    \addplot+[mark=none, color=color3, style = solid, opacity = 0.75] table[x=training_window_length, y=crps_drn, col sep=comma
        ]{Data/crps_training_window_length.txt};
    \addlegendentry{Distributional Regression Network}

    \addplot+[mark=none, color=color4, style = solid, opacity = 0.75] table[x=training_window_length, y=crps_moe, col sep=comma
        ]{Data/crps_training_window_length.txt};
    \addlegendentry{Mixture of Experts}

    \addplot+[mark=none, color=color5, style = solid, opacity = 0.75] table[x=training_window_length, y=crps_qrlt, col sep=comma
        ]{Data/crps_training_window_length.txt};
    \addlegendentry{Quantile Regression}

    \addplot+[mark=none, color=color5, style = solid, opacity = 0.75] table[x=training_window_length, y=crps_qrf, col sep=comma
        ]{Data/crps_training_window_length.txt};
    \addlegendentry{Quantile Regression Forest}

    \addplot+[mark=none, color=color7, style = solid, opacity = 0.75] table[x=training_window_length, y=crps_qrn, col sep=comma
        ]{Data/crps_training_window_length.txt};
    \addlegendentry{Quantile Regression Network}
\end{axis}

\end{tikzpicture}

% Place one global legend below the group
\pgfplotslegendfromname{mylegendconv}

%\end{document}
  \caption{Convergence of the post-processing methods measured via the $CRPS$ as the training set size increases. The curves represent the average across the different lead times of the $CRPS$ given a size of the training set. A strong annual seasonality can be observed for all the methods tested in this work. Non-parametric methods such as Quantile Regression Forest or Quantile Regression Network requires more training data than other methods due to their higher complexity.}
  \label{fig:convergence}
\end{figure}

\clearpage
\section{Appendix E : Influence of the ensemble size on the CRPS}\label{appendixE}
The Continuous Ranked Probability Score is a probabilitic scoring rule used to evaluate probabilistic forecast and especially Ensemble Prediction Systems. However, one could ask how the metric evolve when the EPS evolves and its ensemble size, i.e the number of member in the ensemble, changes. Since the metric evaluates the Cumulative Distribution Function CDF of the ensemble, increasing the number of members will improve the estimated empirical CDF and thus improve the score. Thus, Ferro \textit{et al.} introduces corrected versions of the CRPS called fair-CRPS and adjusted-CRPS that accounts for ensemble size such that the score is consistent with increases in number of members or when comparing forecasts with different ensemble sizes \cite{ferro_2008}.

The fair-CRPS, $fCRPS$, is the estimated CRPS assuming that the ensemble is of an infinite size. The ensemble would be made with infinitely many numbers sampled from the same distributions as the existing members. The adjusted-CRPS, $aCRPS$, is the estimated CRPS assuming an ensemble size of $M$ instead of $m$. Let $y$ be the observation, $\hat{y}$ the predicted ensemble of size $m$ and $F(\hat{y})$ its CDF, then :
\begin{equation}
  fCRPS(F(\hat{y}), y) = CRPS(F(\hat{y}), y) - \int \dfrac{F(x)(1-F(x))}{m-1}dx
\end{equation}

\begin{equation}
  aCRPS(F(\hat{y}), y) = CRPS(F(\hat{y}), y) - \int \dfrac{\left(1-\dfrac{m}{M}\right)F(x)(1-F(x))}{m-1}dx
\end{equation}
with $M$ the desired ensemble size. We can notice than when $M \rightarrow \infty$ then the $aCRPS$ and $fCRPS$ are equal. Those metrics allow to compare forecasts with different ensemble sizes.

In our case our two baselines, the climatological baseline and climatological bootstrap baseline have different ensemble sizes. There are \si{30} members in the climatology and \si{51} in the bootstrap. In figure \ref{fig:CRPS_skill_best} it seems like the climatological boostrap baseline is better than the climatology by a tiny margin. It is in fact an artefact steming from the different ensemble sizes as we can see in figure \ref{fig:fair_crps}. Indeed, the improvements that one can observe on the CRPSS disappears when using the correction accounting for the bigger ensemble in the bootstrap baseline.

\begin{figure}[ht!]
  \centering
  %\documentclass{standalone}
%\usepackage{tikz}
%\usetikzlibrary{patterns}
%\usepackage{pgfplots}
%\pgfplotsset{compat=1.18}
%\usepackage[dvipsnames]{xcolor}

%\definecolor{color0}{HTML}{D51317}
%\definecolor{color1}{HTML}{F39200}
%\definecolor{color2}{HTML}{EFD500}
%\definecolor{color3}{HTML}{95C11F}
%\definecolor{color4}{HTML}{007B3D}
%\definecolor{color5}{HTML}{31B7BC}
%\definecolor{color6}{HTML}{0094CD}
%\definecolor{color7}{HTML}{164194}
%\definecolor{color8}{HTML}{6F286A}

%\begin{document}

\begin{tikzpicture}
    \begin{axis}[
        width=15cm,
        height=10cm,
        xlabel={Lead time [D]},
        xtick={0,4,9,14,19,24,29,34,39,45},
        xticklabels={1, 5, 10, 15, 20, 25, 30, 35, 40, 46},
        ylabel={Probabilistic Skill Score ($CRPSS$, $fCRPSS$, $aCRPSS$)},
        ymin=-0.01,
        ymax=0.011,
        xmin=0, xmax=45,
        grid=major,
        thick,
        mark size=2pt,
        legend columns = 3,
        legend style={at={(0.45,0.01)}, anchor=south west},
        ]

        \addplot [
          draw=none,
          forget plot,
            %fill=gray!30,     % color and transparency
         pattern=north east lines,
        pattern color= black!40,
        ] coordinates {
        (0,-0.01) (6,-0.01) (6,0.011) (0,0.011)};

        \addplot+[ mark=none, line width=1pt, color=brown, style=solid
        ]table[x=lead_time, y=crpss, col sep=comma, header=true]{Data/crps_baselines.txt};
        \addlegendentry{CRPSS}
        
        \addplot+[ mark=none, line width=1pt, color = brown, style = dashed, 
        ]table[x=lead_time, y=fcrpss, col sep=comma, header=true]{Data/crps_baselines.txt};
        \addlegendentry{fCRPSS}

        \addplot+[ mark=none, line width=1pt, color = brown, style = dashdotdotted, 
        ]table[x=lead_time, y=acrpss, col sep=comma, header=true]{Data/crps_baselines.txt};
        \addlegendentry{aCRPSS}

    \end{axis}
\end{tikzpicture}
%\end{document}
  \caption{$CRPSS$, $fCRPSS$ and $aCRPSS$ evolution with the lead time for the climatological bootstrap baseline. The fair and adjusted CRPSS are overlapping for most of the lead time. The lead times between $1$ and $7$ days are hatched to remind that short-term weather forecasts would be more suitable to provide power forecasts at such horizons. For each lead time, the skill scores are computed over all the available samples.}
  \label{fig:fair_crps}
\end{figure}

\clearpage

\end{appendix}

\newpage
\bibliography{references}

\end{document}